\pdfoutput=1

\documentclass[11pt]{article}

\usepackage[]{acl}
\usepackage[frozencache,cachedir=./,outputdir=./]{minted}

\usepackage{times}
\usepackage{latexsym}

\usepackage{hyperref}
\usepackage{url}
\usepackage{amsmath}
\usepackage{amsfonts}
\usepackage{subcaption}
\usepackage{graphicx}
\usepackage{booktabs}
\usepackage{url}
\usepackage{tabularx}
\usepackage{bigstrut}
\usepackage{multirow}
\usepackage{color}
\usepackage{xcolor}
\usepackage{verbatimbox}
\usepackage{enumitem}
\usepackage{xspace}
\usepackage{bbding}
\usepackage{algorithm} 
\usepackage{algpseudocode}
\usepackage{enumitem}
\usepackage{color}
\usepackage{amssymb}%
\usepackage{pifont}%
\newcommand{\cmark}{\ding{51}}%
\newcommand{\xmark}{\ding{55}}%

\usepackage[T1]{fontenc}
\newcommand\mybar{\kern1pt\rule[-\dp\strutbox]{.8pt}{\baselineskip}\kern1pt}
\usepackage{float}
\usepackage{dblfloatfix}
\usepackage[table]{colortbl}
\usepackage{makecell}

\usepackage[T1]{fontenc}

\usepackage[utf8]{inputenc}

\usepackage{microtype}

\usepackage{inconsolata}

\title{LMDX: Language Model-based Document Information \\ Extraction and Localization}

\author{
Vincent Perot$^{1}$\thanks{\ Correspondence to <vperot@google.com>.}, 
Kai Kang$^{2}$,
Florian Luisier$^{2}$,
Guolong Su$^{1}$, \\
\textbf{
Xiaoyu Sun$^{2}$,
Ramya Sree Boppana$^{2}$,
Zilong Wang$^{5}$,
Zifeng Wang$^{3}$, } \\  
\textbf{
Jiaqi Mu$^{1}$,
Hao Zhang$^{4}$,
Chen-Yu Lee$^{3}$,
Nan Hua$^{1}$} \\
$^{1}$Google DeepMind $^{2}$Google Cloud\\ $^{3}$Google Cloud AI Research 
$^{4}$Google $^{5}$UC San Diego \\
}

\begin{document}
\maketitle
\begin{abstract}
Large Language Models (LLM) have revolutionized Natural Language Processing (NLP), improving state-of-the-art and exhibiting emergent capabilities across various tasks. However, their application in extracting information from visually rich documents, which is at the core of many document processing workflows and involving the extraction of key entities from semi-structured documents, has not yet been successful. The main obstacles to adopting LLMs for this task include the absence of layout encoding within LLMs, which is critical for high quality extraction, and the lack of a grounding mechanism to localize the predicted entities within the document. In this paper, we introduce \textit{\textbf{L}anguage \textbf{M}odel-based \textbf{D}ocument Information E\textbf{X}traction and Localization} (LMDX), a methodology to reframe the document information extraction task for a LLM. LMDX enables extraction of singular, repeated, and hierarchical entities, both with and without training data, while providing grounding guarantees and localizing the entities within the document. Finally, we apply LMDX to the PaLM 2-S and Gemini Pro LLMs and evaluate it on VRDU and CORD benchmarks, setting a new state-of-the-art and showing how LMDX enables the creation of high quality, data-efficient parsers.

\end{abstract}

\section{Introduction}
The recent advent of transformers \citep{NIPS2017_3f5ee243} and self-supervised pretraining procedures has led to significant progress in Visually Rich Document (VRD) Understanding. Within that field, the task of document information extraction (IE), which consists of extracting key entities within a semi-structured document (e.g. invoice, tax form, paystub, receipt, etc) given a predefined schema, has received a lot of attention from industry and academia due to its importance and wide applicability to intelligent document processing workflows. However, document information extraction still remains challenging for mainstream systems. 

In particular, information in semi-structured forms is organized in complex layout across many possible templates, which requires deep understanding of the document context, spatial alignment among the different segments of text, and tabular arrangement of hierarchical entities (we define hierarchical entities as entities that are composed of logically grouped leaf entities, e.g. line items on an invoice composed of item description, quantity and price, or deduction items on a paystub composed of deduction type and amount, etc.). 
Moreover, since some business document automation workflows require certain level of accuracy, they are often integrated with human-in-the-loop interactions for auditing and correction of predictions, requiring knowing the precise location of extracted entities to make it a tractable task for a human rater. Finally, since a quasi-infinite number of document types exist, and that organizations have limited annotation resources, most parsers are built with very small amount of training data. From those complexities emerge the following desiderata of document information extraction systems: they should \textbf{(1) support high-quality extraction of singular, repeated, and hierarchical entities}, while \textbf{(2) localizing those entities in the document}, and \textbf{(3) do so with very low or no data annotation cost}. 

So far, no publicly disclosed system has been able to address all of those desiderata. Current mainstream document IE systems are based on sequence-tagging and sequence-generation. Sequence-tagging approaches~\citep{62af8aadfc2c4629ba34f021a17ae9e8, lee2021rope,lee-etal-2022-formnet,lee-etal-2023-formnetv2,wang-etal-2023-queryform} classifies each token into Inside-Outside-Begin (IOB) tags \citep{ramshaw-marcus-1995-text}, which support extraction and localization of leaf entities. However, it is non-trivial to extend these methods for hierarchical entities. Sequence-generation based methods~\citep{10.1007/978-3-030-86331-9_47,kim2022donut} treats extraction as text generation with auto-regressive decoders~\citep{sutskever2014sequence}. Although this line of work allows to predict hierarchical entities, it does not allow localizing entities in the document.
Moreover, both categories require significant human annotation cost to ensure a high quality extraction. 
Thus, a unified IE framework addressing all three desiderata is highly valuable.

In parallel, Large Language Models (LLMs) \citep{openai2023gpt4, palm2, hoffmann2022training} have revolutionized Natural Language Processing, showing the capabilities to solve diverse tasks with an instruction \citep{wei2022finetuned} or a few demonstrations attached to the prompt \citep{brown2020language}. This paradigm shift opens the possibility of extracting entities while addressing all the aforementioned desiderata, but using LLMs for VRD IE has been underexplored. \citet{wang2023docllm} proposes a document-centric LLM and frames IE as a question-answer task, enabling zero-shot extraction but lacking support of localization and hierarchical entity extraction. Furthermore, this can suffer from hallucinations, a common issue with LLMs \citep{huang2023survey}.

\begin{figure*}[ht]
    \centering
    \includegraphics[width=1.0\linewidth]{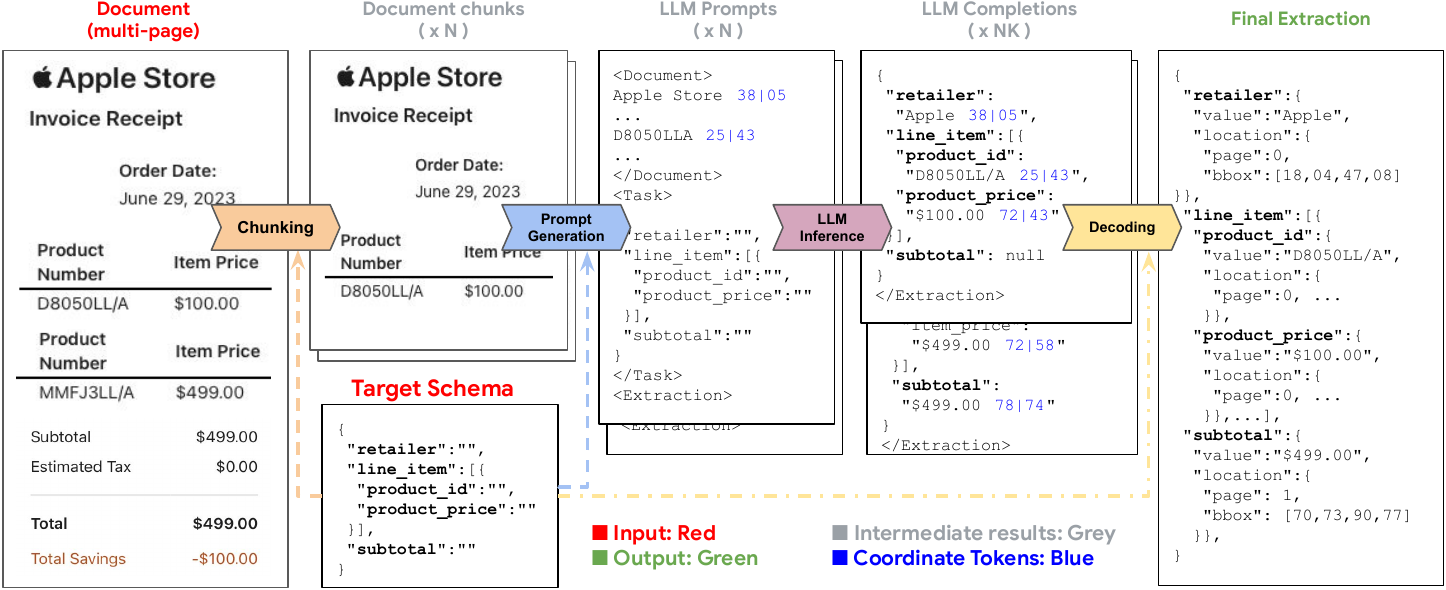}
    \caption{Overview of the LMDX methodology, decomposing the information extraction and localization task in 4 stages in order to frame it for an LLM.
    From the document, we generate LLM prompts containing both the text content and \textit{coordinate tokens} (in \textcolor{blue}{color blue}), which communicates the layout modality (needed for a high-quality extraction) and act as unique identifiers of the text segments. The prompts also contain the target schema, enabling zero-shot information extraction. The LLM completions, in JSON format, naturally support hierarchical entity extraction (e.g. \textit{line\_item}), and include both entity values and segment identifiers, enabling both entity localization (i.e. computing entity bounding box) and removing LLM hallucination through our decoding algorithm.
    }
    \label{fig:system-overview}
\end{figure*}

This motivates us to introduce \textit{\textbf{L}anguage \textbf{M}odel-based \textbf{D}ocument Information E\textbf{X}traction and Localization} (LMDX), a principled methodology for leveraging existing LLMs for information extraction and localization on visually-rich documents, meeting all three identified desiderata of extraction systems and detailed in Figure~\ref{fig:system-overview}.
A comparison of LMDX characteristics and other document information extraction systems can be found at Table~\ref{table:system-comparison}. Our contributions can be summarized as follows:
\begin{itemize}
    \item We present a principled recipe that enables LLMs to perform the document IE task on leaf and hierarchical entities with precise entity localization, including without any training data, and using only the simple text-in, text-out interface that is applicable to LLMs.
    \item We propose a layout encoding scheme that communicates spatial information to the LLM without any changes to its architecture.
    \item We introduce a decoding algorithm that transforms the responses from the LLM into extracted entities and their corresponding bounding boxes on the document, while discarding any LLM hallucination.
    \item We systematically evaluate the data efficiency of LMDX across multiple public benchmarks, establishing a new state-of-the-art, and provide extensive study of the different core designs to demonstrate their effectiveness.
\end{itemize}

\begin{table}[ht]
\caption{Comparison of document information extraction systems. Unlike mainstream document IE systems, LMDX enables the zero-shot extraction, including hierarchical entities, all while localizing its predictions.}
\label{table:system-comparison}
  \centering
  \resizebox{1.0\linewidth}{!}{
    \setlength{\tabcolsep}{0.mm}{
\begin{tabular}{ccccc}
\toprule
\bf Document Information & \bf Hierarchical & \bf Entity & \bf Zero-shot \\ 
\bf Extraction Systems & \bf entity & \bf localization & \bf support \\ 
\cmidrule{1-4}
LayoutLMv3 \citep{huang2022layoutlmv3}, & \multirow{1.75}[2]{*}{\xmark} & \multirow{1.75}[2]{*}{\cmark} & \multirow{1.75}[2]{*}{\xmark} \\
FormNetV2 \citep{lee-etal-2023-formnetv2}, & & & \\
\cmidrule{1-4}
Donut \citep{kim2022donut} &\cmark &\xmark &\xmark \\
\cmidrule{1-4}
DocLLM \citep{wang2023docllm} &\xmark &\xmark &\cmark \\
\cmidrule{1-4}
\textbf{LMDX (Ours)} &\cmark &\cmark &\cmark \\
\bottomrule
\end{tabular}
}}
\end{table}

\vspace{-2mm}
\section{Related Work}

\noindent \textbf{Framing IE from VRD.} Information Extraction from VRD is a complex task that can be framed in a variety of ways. Many approaches divide the problem in two sub-tasks: a text recognition/serialization step, typically achieved by an Optical Character Recognition (OCR) service, followed by a parsing step, which finds the relevant entity values from the recognized text. \citet{Xu2020LayoutLMv2MP, Appalaraju_2021_ICCV} frame this parsing step as Named Entity Recognition (NER), encoding each token with a transformer encoder and classifying each document token into IOB tags, allowing extraction and localization of leaf entities only. 
Other approaches treat extraction as a sequence generation problem. \citet{10.1007/978-3-030-86331-9_47} adds an auto-regressive decoder on top of a text-layout-image encoder, all initialized from T5~\citep{2020t5}. This enables to predict hierarchical entities, but not localize entities in the document.
While LMDX still frames VRD IE as a sequence generation task, our work contrasts with prior work by combining the advantages of the different framings as shown in Table~\ref{table:system-comparison}, supporting hierarchical entities, zero-shot extraction and localizing the entities through the introduction of coordinate tokens.

\noindent \textbf{VRD Representation Learning.} As VRDs contain both textual and visual elements whose spatial position is crucial for their understanding, many works explore custom architectures and pretraining strategies to learn the relation between textual, layout and image modalities \citep{lee-etal-2023-formnetv2, appalaraju2023docformerv2, zhang2022multimodal}. \citet{xu2020layoutlm} uses a separate image encoder before adding the output as feature to the token encodings, while \citet{huang2022layoutlmv3} jointly models the page image patches alongside the tokens, using a word-patch alignment self-supervised pretraining task to learn the connection between the modalities. \citet{DBLP:journals/corr/abs-2108-04539} proposes to encode the relative 2D distances of text blocks in the attention of the transformer, and learning from unlabeled documents with an area-masking strategy. \citet{kim2022donut, lee2023pix2struct} foregoes the text modality completely, using a Vision Transformer encoder with an auto-regressive decoder pretrained on a pseudo-OCR and region masking task on large document image corpora. 
Unlike prior work, LMDX encodes the layout modality solely through text coordinate tokens, hence allows reusing LLMs with no architecture change and foregoing expensive vision encoders, while achieving state-of-the-art results.

\vspace{-3mm}
\noindent \textbf{LLMs for Extraction} has mostly been studied in the text domain \citep{keraghel2024survey}, either generally \citep{laskar2023systematic} or domain-specific \cite{de-toni-etal-2022-entities,hu2024improving}. \citet{wang2023gptner} uses a LLM to insert special tokens to mark the boundaries of target entities. \citet{ashok2023promptner} proposes a NER framework with in-context learning demonstrations, prompting the LLM to output an entities list with explanations justifying its matches with the provided entity definitions.
Yet, LLMs remain underexplored for IE on VRDs. \citet{wang2023docllm} uses unlabeled document corpora and turns existing labeled VRD understanding datasets in instruction tuning format, building a layout-aware LLM with various document understanding capabilities. In contrast, LMDX focuses on IE specifically, with an emphasis on hierarchical entity and entity localization support.

\section{LMDX Methodology}
\label{sec:methodology}
\vspace{-0.5mm}

Overall, our pipeline is divided into four stages: chunking, prompt generation, LLM inference and decoding, detailed in the following sections. An overview with a simple example can be found in Figure~\ref{fig:system-overview}, with the input and output of each stage showcased. 
In this example, the target extraction schema contains two leaf entity types \textit{retailer} and \textit{subtotal}, and one hierarchical entity type \textit{line\_item}, composed of a \textit{product\_id} and a \textit{product\_price}.

\noindent \textbf{Input Document.} The input to our pipeline is the document's text segments (lines and words) along with their corresponding spatial position (bounding box) on the pages, typically obtained with an OCR service or a PDF rendering engine.

\subsection{First Stage: Chunking}

While some LLMs support long context (hundreds of thousands of tokens), not all LLMs can fit the entire document within its prompt, as documents can be hundreds of pages long. Thus, the document is divided into document chunks so that each is small enough to be processed by the LLM. To achieve this, we first divide the document into individual pages, then we iteratively remove the last line segments until the prompt containing this chunk is below the maximum input token length of the LLM. Lastly, we group those removed lines as a new document page, and repeat the same logic until all chunks are below the input token limit of the LLM. At the end of this stage, we have $N$ chunks.
The decision to first divide the document by page stems from the observation that entities rarely cross page boundaries, and as such this chunking scheme will have minimal impact on the final extraction quality. The algorithm is described in pseudo-code in Appendix~\ref{appendix:chunking}.

\subsection{Second Stage: Prompt Generation}
\label{sec:prompt-generation}

\begin{figure}[ht]
\begin{center}
\includegraphics[width=0.6\linewidth]{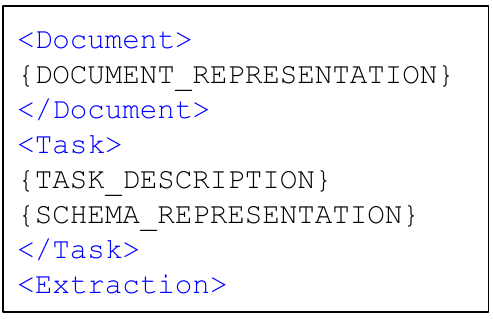}
\end{center}
\vspace{-2mm}
\caption{Structure of the LLM prompts.}
\label{fig:prompt-design}
\end{figure}

The prompt generation stage takes in the $N$ document chunks and creates a LLM prompt for each of them. As seen in Figure~\ref{fig:prompt-design}, our prompt design contains the document representation, a description of the task, and the target schema representation containing the entities to extract. XML-like tags are used to define the start and end of each component.

\noindent \textbf{Document Representation.} The chunk content is represented in the prompt as the concatenation of all its segment texts (lines or words from OCR), suffixed with the coordinates of those segments (derived from the bounding boxes) in the following format:
$\texttt{<}\texttt{segment~text}\texttt{>}~\texttt{XX|YY}_\texttt{segment}$.~Coordinate tokens, $\texttt{XX}$ and $\texttt{YY}$, are built by normalizing the segment's X and Y coordinates, and quantizing them in $B$ buckets, assigning the index of that bucket as the token for that coordinate.

Encoding the coordinates as tokens within the prompt allows us to communicate the layout modality to the LLM, without any change to its architecture. There are many variations to that scheme: using line versus words as segment, the granularity of the quantization, and the number of coordinates to use per segment (e.g. $[x\textsubscript{center}, y\textsubscript{center}]$ versus $[x\textsubscript{min}, y\textsubscript{min}, x\textsubscript{max}, y\textsubscript{max}]$).
Appendix~\ref{appendix:token-length-stats} shows how those variations affect the prompt token length. Experimentally, we've found using line-level segments with 2 coordinates $[x\textsubscript{center}, y\textsubscript{center}]$ and $B = 100$ quantization buckets worked best, as detailed in Appendix~\ref{appendix:coordinate_schemes}. Hence, we've adopted that coordinate tokenization scheme in our experiments.

\noindent \textbf{Task Description.} The task description is simply a short explanation of the task to accomplish. In our experiments, we set it to the following: \textit{From the document, extract the text values and tags of the following entities:}.

\noindent \textbf{Schema Representation.} The schema is represented as a structured JSON object, where the keys are the entity types to be extracted, and the values correspond to their occurrence (single or multiple) and sub-entities (for hierarchical entities). For instance, \textit{\{"foo": "", "bar": [\{"baz": []\}]\}} means that the LLM should extract only a single entity of type \textit{foo} and multiple hierarchical entities of type \textit{bar}, that could each hold multiple entities of type \textit{baz}.

After this step, we have $N$ prompts, one for each document chunk. An example of a prompt on a document can be found in Appendix~\ref{appendix:sample_prompts_and_completions}, Figure~\ref{fig:vrdu_adbuy_sample}.

\subsection{Completion Targets}
\label{sec:completion-targets}

In this section, we describe the expected LLM completion format, which can be observed in \textit{LLM Completions} section of  Figure~\ref{fig:system-overview}. Like the schema, the completion is a JSON structured object with the keys being the entity types, and values being the extracted information from the document chunk. JSON was chosen as a format for the completion and schema since it supports hierarchical objects (hence hierarchical entities), is very token-efficient, and emitting JSON is within mainstream LLMs' capabilities \citep{jsonformer,openaijson}. Note that the keys in the completion have the same ordering, occurrence and class (hierarchical or leaf) as the entity types in the schema. The values of leaf entities must follow a specific format:

\begin{center}
$\texttt{<}\texttt{text~on~segment}_\texttt{1}\texttt{>}~\texttt{XX|YY}_\texttt{segment1} \texttt{\symbol{92}n}$
$\texttt{<}\texttt{text~on~segment}_\texttt{2}\texttt{>}~\texttt{XX|YY}_\texttt{segment2} \texttt{\symbol{92}n}~\texttt{...}$
\end{center}

An entity can span multiple (potentially disjoint) text segments (lines or words). For each segment of the entity, the value contains the entity text on that segment, along with the coordinate tokens of that segment, which act as a \textit{segment identifier}, uniquely identifying the segment, and allowing us to localize the entities and ground the model prediction (e.g. making sure the extracted value is not a hallucination), as will be detailed in Section~\ref{sec:decoding}. Finally, missing entity types are explicitly completed by the model with $null$ for singular types, and $[]$ for repeated types. Samples of completions can be found in Appendix~\ref{appendix:sample_prompts_and_completions}, Figure~\ref{fig:vrdu_adbuy_sample}.

\subsection{Third Stage: LLM Inference}
\label{sec:llm-inference}

In this stage of the pipeline, we run inference on the LLM with the $N$ prompts. For each prompt, we sample $K$ completions from the LLM (for a total of $NK$ completions for the entire document) using Top\textsubscript{K} sampling. This randomness in the sampling allows to do error correction (e.g. if a response is not valid JSON, have hallucinated segment coordinate identifier, etc), and increase the extraction quality as will be shown in Section~\ref{sec:ablation}. We use a fixed random seed to get a deterministic inference.

\subsection{Fourth Stage: Decoding}
\label{sec:decoding}

In this stage (\textit{Decoding} in Figure~\ref{fig:system-overview}), we parse the LLM completions into entities and their locations.

\noindent \textbf{Conversion to structured entities.} We begin by parsing each LLM completion as a JSON object. Completions that fail to parse are discarded. For each key-value pair in the JSON object, we interpret the key as the entity type and parse the value to get the entity text and bounding box (as detailed in the next paragraph). Predicted entity types that are not in the target extraction schema are discarded. If the model unexpectedly predicts multiple values for single-occurrence entity types, we use the most frequent value as the final predicted value. Hierarchical JSON objects are recursively parsed as hierarchical entities in a similar manner. This algorithm is described in pseudo-code in Appendix~\ref{appendix:decoding}.

\noindent \textbf{Entity Value Parsing.} 
We expect the JSON value to include both text extractions and segment identifiers for each predicted entity, as described in Section~\ref{sec:completion-targets}. We first parse the value into its (\textit{segment~text}, \textit{segment~identifier}) pairs. For each pair, we look up the corresponding segment in the original document using the segment identifier and verify that the extracted text is \textit{exactly} included on that segment. The entity is discarded if that verification fails, ensuring LMDX discards all LLM hallucinations. Finally, once we have the entity location on all its segments, we get the entity bounding box by computing the smallest bounding box encompassing all the words included in the entity. Entity values with any segments that fail to ground (invalid entity value format, non-existent segment identifier, or non-matching segment text) in the original document are discarded. The entity value parsing algorithm is described in pseudo-code in Appendix~\ref{appendix:value_parsing}, and parsing errors rates are detailed in Appendix~\ref{appendix:parsing_rates}.

\noindent \textbf{Prediction Merging.}
We first merge the predicted entities for the same document chunk from the $K$ LLM completions through majority voting \citep{wang2022self}. For each entity type, we gather the predicted entities, including empty predictions, across the $K$ completions. The most common prediction(s) are selected as the predicted value for that entity type. We then merge the predictions among the $N$ document chunks by concatenating them to obtain the document level predictions.

\noindent \textbf{Prediction Merging for hierarchical entities.} 
For hierarchical entities, we use the entire predicted tree value from a single LLM completion, as this method best preserves the parent-child relationship predicted by the model. For each top-level hierarchical entity type, we perform majority voting on all affiliated leaf, intermediate and top-level entity types among $K$ completions as if they are flattened. We then equally tally the votes to determine which completion to use for the prediction, and select the most common one for that hierarchical entity.

\section{Evaluation}
\vspace{-0.5mm}

We seek to evaluate the effectiveness of LMDX on public IE benchmarks, and apply it to two distinct LLMs to validate the generality of the methodology: PaLM 2-S \citep{palm2} and Gemini Pro \citep{DBLP:journals/corr/abs-2312-11805}, which we call LMDX\textsubscript{PaLM 2-S} and LMDX\textsubscript{Gemini Pro} respectively.

Firstly, starting from their original checkpoint, we finetune those LLMs on the prompts and completions detailed in Section~\ref{sec:prompt-generation} and \ref{sec:completion-targets} on a data mixture containing a variety of \textit{(document, schema, extraction)} tuples. In particular, this data mixture contains the \textit{Payment} dataset \citep{majumder2020representation}, along with a diverse set of publicly available PDF form templates obtained from government websites that we filled with synthetic data using an internal tool, and annotated for schema and entities to extract. The goal of this tuning is to obtain a \textit{Base Entity Extractor} checkpoint by training the model to learn the IE task along with our desired extraction syntax.
No document or schema contained in the base extraction training phase overlap with the documents and schemas used in our target benchmarks, hence we use those LLMs for zero-shot information extraction evaluation on the target benchmarks.

\noindent \textbf{Finetuned Performance.} We are also interested in evaluating how data-efficient LMDX is (e.g. how quickly it learns information extraction on a new document type). To answer this, starting from the \textit{Base Entity Extractor} checkpoint, we finetune the LLM directly on the target benchmark.

\noindent \textbf{Parameters}. For training, we finetune using a batch size of 8, a dropout probability of 0.1 and a learning rate of $10^{-6}$ with a standard cross-entropy loss for 4000 steps on TPUv4~\citep{jouppi2023tpu}. Once training is done, for finetuned experiments, we select the checkpoint with the lowest loss on the dev set, and report performance on the test set. For LLM inference, we use a temperature of 0.5 and a Top\textsubscript{K} of 40, sampling 16 responses for each chunk processed by the LLM, as described in Section \ref{sec:llm-inference}. Finally, for both training and inference, we use an input token length of 6144 and output token length of 2048. We use line-level segments and only two coordinates $[x\textsubscript{center}, y\textsubscript{center}]$ with 100 quantization buckets, as supported by Appendix~\ref{appendix:coordinate_schemes}.

\subsection{Benchmarks}

\noindent \textbf{Visually Rich Document Understanding (VRDU).} \citet{10.1145/3580305.3599929} introduces two public visually-rich documents IE benchmarks: \textit{Registration Form}, containing 6 semantically rich entity types, and \textit{Ad-buy Form}, containing 14 entity types with one hierarchical \textit{line\_item} entity type. For each benchmark, VRDU proposes samples of 10, 50, 100 and 200 train documents with high-quality OCR\footnote{https://cloud.google.com/vision/docs/ocr} which we use to evaluate the data efficiency of LMDX. It also offers different tasks to evaluate the generalization powers of extraction systems: Single Template Learning (STL) where train/test splits share the same single template, Unseen Template Learning (UTL) where train/test contain disjoint sets of templates, and Mixed Template Learning (MTL) where train/test contain overlapping sets of templates. We report Micro-F1 through the provided evaluation tool. For VRDU, we only run the finetuning experiments using LMDX\textsubscript{PaLM 2-S}, given the significant cost of finetuning on all its tasks and train split sizes.

\noindent \textbf{Consolidated Receipt Dataset (CORD).}\footnote{https://huggingface.co/datasets/naver-clova-ix/cord-v1} \citet{park2019cord} introduces a benchmark of Indonesian receipts from shops and restaurants, with a target schema of 30 fine-grained entities, grouped into \textit{menu}, \textit{total} and \textit{subtotal} hierarchical entities. We adopt the evaluation tool from prior work \citep{kim2022donut} and report Micro-F1 on that benchmark. For our experiments, we use the official $train$ ($|\mathcal{D}|=800$), $dev$ and $test$ splits. To evaluate LMDX's data efficiency, we further sample the first $|\mathcal{D}|=10/50/100/200$ documents from the $train$ split. For each of those data setup, we finetune LMDX for 12000 steps. For comparison, we also train and evaluate state-of-the-art baselines $\rm{LayoutLMv3_{LARGE}}$ and $\rm{Donut}$. Those baselines are detailed in Appendix~\ref{appendix:cord_baseline}.

\noindent \textbf{Baselines} In the zero-shot setting, we compare LMDX to other LLM baselines: GPT-3.5, Gemini Pro \citep{DBLP:journals/corr/abs-2312-11805}, and PaLM 2-S \citep{palm2} that we prompt with the raw OCR text and IE instruction (called \textbf{GPT-3.5\textsubscript{+OCR}}, \textbf{Gemini Pro\textsubscript{+OCR}} and \textbf{PaLM 2-S\textsubscript{+OCR}} respectively). We also compare LMDX with strong Vision-Language models: LLaVA-v1.5-13B \citep{liu2023improved}, Gemini Pro, and GPT-4V \citep{openai2023gpt4} that we prompt with the document page image and IE instructions (called \textbf{LLaVA-v1.5-13B\textsubscript{+Image}}, \textbf{GPT-4V\textsubscript{+Image}}, and \textbf{Gemini Pro\textsubscript{+Image}}). Those baselines are fully detailed in Appendix~\ref{appendix:common_baseline}. Unlike LMDX, those large model baselines do not localize their predictions.
In the finetuned setting, we compare LMDX\textsubscript{PaLM 2-S} to popular VRD IE baselines. For VRDU, we compare to its published baselines \cite{10.1145/3580305.3599929}, $\rm{LayoutLM/v2/v3}$ and $\rm{FormNet}$. For CORD, we train and evaluate state-of-the-art baselines $\rm{LayoutLMv3}$ and $\rm{Donut}$. Those baselines are detailed in Appendix~\ref{appendix:cord_baseline}.
Following prior work \citep{lee-etal-2022-formnet, lee-etal-2023-formnetv2, xu2020layoutlm, Xu2020LayoutLMv2MP}, for all models leveraging the text modality, we use the benchmarks' provided OCR, ensuring a fair comparison.

\subsection{Results}

\begin{table*}[ht]
  \caption{Results of LMDX\textsubscript{PaLM 2-S} and LMDX\textsubscript{Gemini Pro} on the different tasks and training data size setups $|\mathcal{D}|$ of VRDU, with best and second best performing model results in bold and underlined respectively, with Micro-F1 reported. We specify the modalities leveraged by each model ($T \rightarrow Text$, $L \rightarrow Layout$, $I \rightarrow Image$) and whether their entities are localized.}
  \label{table:vrdu-results}%
  \centering
  \small
  \resizebox{\linewidth}{!}{
    \setlength{\tabcolsep}{2mm}{
\begin{tabular}{cccc|ccc|ccc}
\toprule
\multirow{3}[6]{*}{\textbf{$|\mathcal{D}|$}} & \multirow{3}[6]{*}{\textbf{Model}} & \multicolumn{1}{c}{\multirow{3}[6]{*}{\rotatebox[origin=c]{90}{\phantom{||}\textbf{Modality}}}} & \multicolumn{1}{c}{\multirow{3}[6]{*}{\rotatebox[origin=c]{90}{\phantom{|||||||||}\textbf{Localized?}}}} & \multicolumn{3}{c}{\textbf{Registration Form}} & \multicolumn{3}{c}{\textbf{Ad-buy Form}} \\
\cmidrule(lr){5-7} \cmidrule(lr){8-10}
      &  &  & \multicolumn{1}{c}{}  & \multicolumn{1}{c}{\textbf{Single}} & \multicolumn{1}{c}{\textbf{Unseen}} & \multicolumn{1}{c}{\textbf{Mixed Template}} &  \multicolumn{1}{c}{\textbf{Unseen}} & \multicolumn{2}{c}{\textbf{Mixed Template}} \\
\cmidrule(lr){5-5} \cmidrule(lr){6-6} \cmidrule(lr){7-7} \cmidrule(lr){8-8}  \cmidrule(lr){9-10}
&   &  &  \multicolumn{1}{c}{} & \multicolumn{1}{c}{\multirow{1}[1]{*}{\textbf{Micro-F1}}} & \multicolumn{1}{c}{\multirow{1}[1]{*}{\textbf{Micro-F1}}} & \multicolumn{1}{c}{\multirow{1}[1]{*}{\textbf{Micro-F1}}} &  \multicolumn{1}{c}{\multirow{1}[1]{*}{\textbf{Micro-F1}}} & \multicolumn{1}{c}{\multirow{1}[1]{*}{\textbf{Micro-F1}}} & \multicolumn{1}{c}{\multirow{1}[1]{*}{\textbf{Line Item F1}}} \\
\midrule
\multirow{7.75}[2]{*}{\textbf{0}} & LLaVA-v1.5-13B\textsubscript{+Image} & \textit{I} & \xmark & 5.29 & 5.05 & 5.00 & 0.38 & 0.34 & 0.00 \\
& GPT-4V\textsubscript{+Image} & \textit{I} & \xmark & 68.97 & 69.44 & 65.34 & 31.84 & 31.95 & 4.45 \\ 
& Gemini Pro\textsubscript{+Image} & \textit{I} & \xmark & 53.90 & 53.72 & 48.60 & 15.24 & 15.38 & 0.91 \\ 
& Gemini Pro\textsubscript{+OCR} & \textit{T} & \xmark & 73.62 & 73.66 & 69.41 & 32.90 & 34.46 & 19.25 \\ 
& PaLM 2-S\textsubscript{+OCR} & \textit{T} & \xmark & 62.80 & 63.51 & 59.78 & 29.70 & 30.24 & 9.86 \\ 
& GPT-3.5\textsubscript{+OCR} & \textit{T} & \xmark & 67.23 & 67.49 & 63.86 & 29.84 & 30.05 & 7.65 \\ 
& \textbf{LMDX\textsubscript{PaLM 2-S}} & \textit{T+L} & \cmark & \underline{73.81} &  \underline{74.94} & \underline{71.65} & \textbf{39.33} & \textbf{39.74} & \underline{21.21} \\
& \textbf{LMDX\textsubscript{Gemini Pro}} & \textit{T+L} & \cmark & \textbf{76.78} &  \textbf{77.18} & \textbf{75.15} & \underline{37.94} & \underline{38.02} & \textbf{23.29} \\
\midrule
\multirow{5.}[2]{*}{\textbf{10}} & {FormNet} & \textit{T+L} & \cmark & 74.22 & 50.53 & 63.61 & 20.28  & 20.47 & 5.72 \\
      & {LayoutLM} & \textit{T+L+I} & \cmark & 65.91 & 25.54 & 36.41 & 19.92 & 20.20 & 6.95\\
      & {LayoutLMv2} & \textit{T+L+I} & \cmark & \underline{80.05} & \underline{54.21} & \underline{69.44} & \underline{25.17} & \underline{25.36} & \underline{9.96}\\
      & {LayoutLMv3} & \textit{T+L+I} & \cmark & 72.51 & 21.17 & 60.72 & 10.01 & 10.16 & 5.92 \\
      & \textbf{LMDX\textsubscript{PaLM 2-S}} & \textit{T+L} & \cmark & \textbf{90.88} & \textbf{86.87} & \textbf{87.72} & \textbf{54.82} & \textbf{54.35} & \textbf{39.35} \\
\midrule
\multirow{5.}[2]{*}{\textbf{50}} & {FormNet} & \textit{T+L} &\cmark & \underline{89.38} & \underline{68.29} & \underline{85.38} & 39.52 & 40.68 & 19.06 \\
      & {LayoutLM} & \textit{T+L+I} & \cmark & 86.21 & 55.86 & 80.15 & 38.42 & 39.76 & 19.50 \\
      & {LayoutLMv2} & \textit{T+L+I} & \cmark & 88.68 & 61.36 & 84.13 & \underline{41.59} & \underline{42.23} & \underline{20.98} \\
      & {LayoutLMv3} & \textit{T+L+I} & \cmark & 87.24 & 47.85 & 81.36 & 38.43 & 39.49 & 19.53 \\
      & \textbf{LMDX\textsubscript{PaLM 2-S}} & \textit{T+L} & \cmark & \textbf{93.06} & \textbf{88.43} & \textbf{91.42} & \textbf{75.70} & \textbf{75.08} & \textbf{65.42} \\
\midrule
\multirow{5.}[2]{*}{\textbf{100}} & {FormNet} & \textit{T+L} & \cmark & \underline{90.91} & \underline{72.58} & 88.13 & 39.88 & 40.38 & 18.80 \\
      & {LayoutLM} & \textit{T+L+I} & \cmark & 88.70 & 63.68 & 86.02 & 41.46 & 42.38 & 21.26 \\
      & {LayoutLMv2} & \textit{T+L+I} & \cmark & 90.45 & 65.96 & \underline{88.36} & \underline{44.35} & \underline{44.97} & \underline{23.52} \\
      & {LayoutLMv3} & \textit{T+L+I} & \cmark & 89.23 & 57.69 & 87.32 & 41.54 & 42.63 & 22.08 \\
      & \textbf{LMDX\textsubscript{PaLM 2-S}} & \textit{T+L} & \cmark & \textbf{93.97} & \textbf{89.70} & \textbf{92.41} & \textbf{75.99} & \textbf{78.05} & \textbf{69.77} \\
\midrule
\multirow{5.}[2]{*}{\textbf{200}} & {FormNet} & \textit{T+L} & \cmark & \underline{92.12} & \underline{77.29} & \underline{90.51} & 42.87 & 43.23 & 21.86 \\
      & {LayoutLM} & \textit{T+L+I} & \cmark & 90.47 & 70.47 & 87.94 & 44.18 & 44.66 & 23.90 \\
      & {LayoutLMv2} & \textit{T+L+I} & \cmark & 91.41 & 72.03 & 89.19 & \underline{46.31} & \underline{46.54} & \underline{25.46}\\
      & {LayoutLMv3} & \textit{T+L+I} & \cmark & 90.89 & 62.58 & 89.77 & 44.43 & 45.16 & 24.51 \\
      & \textbf{LMDX\textsubscript{PaLM 2-S}} & \textit{T+L} & \cmark & \textbf{93.97} & \textbf{90.22} & \textbf{92.78} &  \textbf{78.42} & \textbf{79.82} & \textbf{72.09} \\
\bottomrule
\end{tabular}
    }
    }
\end{table*}

\begin{table*}[h!b]
  \caption{Results of LMDX\textsubscript{PaLM 2-S} and LMDX\textsubscript{Gemini Pro} on the different training data size setups $|\mathcal{D}|$ of CORD, compared to zero-shot (Large Models) and finetuned ($\rm{LayoutLMv3}$ and $\rm{Donut}$) baselines. Micro-F1 is reported, with best and second best performing model results in bold and underlined respectively.}
  \label{table:cord-result}
  \centering
  \small
  \resizebox{1.0\linewidth}{!}{
    \setlength{\tabcolsep}{2.5mm}{
\begin{tabular}{ccccccccc}
\toprule
\textbf{Model} & \textbf{Modality} &  \textbf{Localized?} & \multicolumn{1}{c}{\multirow{0.75}[1]{*}{$|D|$=0}} & \multicolumn{1}{c}{\multirow{0.75}[1]{*}{$|D|$=10}} & \multicolumn{1}{c}{\multirow{0.75}[1]{*}{$|D|$=50}} & \multicolumn{1}{c}{\multirow{0.75}[1]{*}{$|D|$=100}} & \multicolumn{1}{c}{\multirow{0.75}[1]{*}{$|D|$=200}} & \multicolumn{1}{c}{\multirow{0.75}[1]{*}{$|D|$=800}}\\

\midrule
LLaVA-v1.5-13B & \textit{I} & \xmark & 5.97 & - & - & - & - & - \\
GPT-4V\textsubscript{+Image} & \textit{I} & \xmark & 64.05 & - & - & - & - & - \\ 
Gemini Pro\textsubscript{+Image} & \textit{I} & \xmark & 47.12 & - & - & - & - & - \\ 
Gemini Pro\textsubscript{+OCR} & \textit{T} & \xmark & 59.57 & - & - & - & - & - \\ 
PaLM 2-S\textsubscript{+OCR} & \textit{T} & \xmark & 55.85 & - & - & - & - & - \\ 
GPT-3.5\textsubscript{+OCR} & \textit{T} & \xmark & 48.92 & - & - & - & -  & - \\ 
 {Donut} & \textit{I} & \xmark & 0.00 & 26.15 & 65.68 & 71.81 & 75.85 & 81.55 \\
 {LayoutLMv3} & \textit{T+L+I} & \cmark & 0.00 & 74.04 & 85.78 & 90.39 & \underline{93.59} & \textbf{95.66} \\
 \textbf{LMDX\textsubscript{PaLM 2-S}} & \textit{T+L} & \cmark & \textbf{66.95} & \textbf{90.02} & \underline{91.40} & \textbf{91.48} & 93.40 & 94.51 \\
 \textbf{LMDX\textsubscript{Gemini Pro}} & \textit{T+L} & \cmark & \underline{66.03} & \underline{89.45} & \textbf{91.66} & \underline{91.16} & \textbf{93.76} & \underline{95.57} \\
\bottomrule
\end{tabular}%
}}
\end{table*}%

VRDU results are presented in Table~\ref{table:vrdu-results}. In zero-shot ($|\mathcal{D}|=0$) setting, LMDX\textsubscript{PaLM 2-S} and LMDX\textsubscript{Gemini Pro} have higher extraction quality than all other large models baselines, including the ones using the same LLM and OCR or image, showing improvements brought by the LMDX methodology itself. 
In finetuned setting on VRDU, LMDX\textsubscript{PaLM 2-S} is much more data efficient than the baselines: it is at 5.06\% Micro-F1 of its peak performance at 10 training documents for Registration Form Mixed Template (87.72\% vs 92.78\% Micro-F1) while LayoutLMv2, the strongest finetuned baseline, is within 19.75\% of its peak performance (69.44\% vs 89.19\% Micro-F1), showcasing that it learns extraction on a new document type much faster. Moreover, LMDX\textsubscript{PaLM 2-S} generalizes better to unseen templates than finetuned baselines: on Registration Form, LMDX\textsubscript{PaLM 2-S} has a drop less than 5\% Micro-F1 on Unseen Template compared to Single Template across data regimes, while LayoutLMv2 see a drop between 19\% and 27\%.

\vspace{-0.565mm}

On CORD (in Table~\ref{table:cord-result}), we observe similar trends, reaching state-of-the-art on all but one data regime, highlighting the generality of the results.

\noindent \textbf{Performance on Hierarchical Entities.} To showcase extraction quality on hierarchical entities, we display in Table~\ref{table:vrdu-results} the F1 score on Ad-buy Form Mixed's \textit{line\_item} entity type. Overall, LMDX has much higher Line Item F1 than the baselines for all data regimes. In particular, LMDX\textsubscript{PaLM 2-S} has similar Line Item F1 at zero-shot than the best finetuned baseline at 200 train documents (21.21\% versus 25.46\% respectively). With all the training data, LMDX\textsubscript{PaLM 2-S} scores a 72.09\% Line Item F1, an absolute improvement of 46.63\% over the best baseline LayoutLMv2. Finally, as LMDX encodes the layout modality, it possesses much higher zero-shot Line Item F1 than large models baselines.

\vspace{2.3mm}
\noindent \textbf{Entity Localization Accuracy.} In order to evaluate the localization quality independently of the extraction quality, we compute the Localization Accuracy of LMDX and all baselines that can localize entities using the formula: $Accuracy_{Localization} = \frac{N_{E+L}}{N_{E}}$ where $N_{E+L}$ is the number of entities correctly extracted and localized, and $N_{E}$ is the number of entities correctly extracted. Since LMDX localizes at the line level, localization verification is done at the line-level as well, i.e. localization is considered correct if the prediction bounding box is covered by the groundtruth line-level bounding box by more than 80\%. We present the results in Table~\ref{table:localization_accuracy}. Overall, LMDX\textsubscript{PaLM 2-S} and LMDX\textsubscript{Gemini Pro} can localize their predictions reliably at the line-level with the segment identifiers, with 88\%-94\% accuracy at zero-shot, and 98\%-99\% in finetuned cases, which is slightly higher than baselines.

\begin{table}[t!b]
  \caption{Entity Localization Accuracy on Registration Form and Ad-Buy Form Mixed Benchmarks, for models supporting localization. Best result is in bold.}
  \vspace{-0.2mm}
  \label{table:localization_accuracy}
  \centering
  \small
  \resizebox{\linewidth}{!}{
    \setlength{\tabcolsep}{0mm}{
\begin{tabular}{cc||ccccc}
\toprule
\phantom{||||||||||||} & \textbf{Model} & $|\mathcal{D}|$=0\phantom{||} & $|\mathcal{D}|$=10\phantom{||} & $|\mathcal{D}|$=50\phantom{||} & $|\mathcal{D}|$=100\phantom{||} & $|\mathcal{D}|$=200\phantom{||} \\

\midrule
\multirow{4.5}[5]{*}{\rotatebox[origin=c]{90}{\phantom{||}\textbf{Reg. Form}}} & LayoutLM & N/A & 98.71 & 99.69 & 99.63 & 99.69 \\
 & LayoutLMv2 & N/A & 99.00 & 99.54 & 99.72 & 99.75 \\
 & LayoutLMv3 & N/A & 99.20 & 99.39 & 99.72 & 99.67 \\
 & \textbf{LMDX\textsubscript{PaLM 2-S}}  & 93.21 & \textbf{99.75} & \textbf{99.87} & \textbf{99.92} & \textbf{99.87} \\
 & \textbf{LMDX\textsubscript{Gemini Pro}\phantom{||}} & \textbf{94.43} & \textbf{-} & \textbf{-} & \textbf{-} & \textbf{-} \\
\midrule
\multirow{-0.125}[5]{*}{\rotatebox[origin=c]{90}{\phantom{|||}\textbf{Ad-buy Form}}} & LayoutLM & N/A & 92.60 & 95.24 & 95.09 & 95.38 \\
 & LayoutLMv2 & N/A & 93.95 & 95.64 & 95.72 & 95.78 \\
 & LayoutLMv3 & N/A & 90.68 & 95.28 & 95.88 & 95.95 \\
 & \textbf{LMDX\textsubscript{PaLM 2-S}}  & 88.18 & \textbf{94.51} & \textbf{98.28} & \textbf{98.69} & \textbf{98.65} \\
 & \textbf{LMDX\textsubscript{Gemini Pro}\phantom{||}} & \textbf{92.51} & \textbf{-} & \textbf{-} & \textbf{-} & \textbf{-} \\
\bottomrule
\end{tabular}%
}}
\end{table}%

\subsection{Ablation Study}
\label{sec:ablation}

In this section, we ablate different facets of the LMDX methodology to highlight their importance. The results can be found in Table~\ref{table:ablations}. For all ablations, we evaluate LMDX\textsubscript{PaLM 2-S} on the VRDU Ad-Buy Form Mixed Template task at $|\mathcal{D}|=10$ data size, only changing the ablated facet.

\vspace{2.3mm}
\noindent \textbf{Effects of Base Entity Extraction Training.} In this ablation, we remove the initial training on the varied data mixture and directly finetune on the VRDU target task. As seen in Table~\ref{table:ablations}, skipping that training leads to -11.44\% micro-F1 as the model has to learn from scratch the task, the desired completion format and coordinate tokens' semantics.

\vspace{2.3mm}
\noindent \textbf{Effects of Coordinate Tokens.} In this ablation, we replace the coordinate tokens, which communicate the position of each line within the document, by the index of that line. This index still acts as a unique identifier for the line segment (required for entity localization) but does not communicate any position information. An example of a prompt with line index can be found in Appendix~\ref{appendix:sample_prompts_and_completions} Figure~\ref{fig:vrdu_adbuy_sample_line_index}, and per-entity F1 can be found in Appendix~\ref{appendix:per-entity-f1} Table~\ref{table:per-entity-f1}. As seen in Table~\ref{table:ablations}, the coordinate tokens are crucial to quality, leading +14.98\% micro-F1.

\vspace{2.3mm}
\noindent \textbf{Effects of Sampling Strategy.} In this ablation, we discard our strategy of sampling $K=16$ completions per chunk, and instead sample a single response. As seen in Table~\ref{table:ablations}, this leads to a 1.5\% drop in micro-F1. While overall minor for quality, the sampling strategy corrects extraction format mistakes (see parsing error rates in Appendix~\ref{appendix:parsing_rates}), leading to a successful extraction on all documents.

\vspace{2.3mm}
\noindent \textbf{Effects of Missing Entity Types.} In this ablation, we study the effect of having the model's completions skip missing entity types in the completions instead of explicitly outputting $"type": null$ for those (See example in Appendix~\ref{appendix:sample_prompts_and_completions}, Figure~\ref{fig:vrdu_adbuy_sample_missing_types}). As seen in Table~\ref{table:ablations}, this leads to a 6.77\% drop in micro-F1 over explicitly outputting missing types. We hypothesize that this is due to the fact that having completions skip missing types means that, during response generation, the model has to choose with a single token computation budget within the $N$ remaining entity types which one is the next present (essentially a $N$-way classification). %
Explicitly emitting missing entity types means the model only has to copy the types directly from the schema in the prompt, and has to do 2-way classification within a single token computation budget to declare if an entity is present or not (e.g. emit token $null$ if entity is missing or $"$ if present), which is an easier task.

\begin{table}[b!]
  \caption{Ablations of LMDX's core designs. Ablations are done on VRDU Ad-Buy Mixed Template with LMDX\textsubscript{PaLM 2-S} at $|\mathcal{D}|=10$ data size. Micro-F1 is reported. Overall, all components contribute to the final performance.}
  \label{table:ablations}%
  \centering
  \small
  \resizebox{\linewidth}{!}{
    \setlength{\tabcolsep}{1.75mm}
    {
\begin{tabular}{c||c|c|c|c}
\toprule
 \textbf{LMDX} & \textbf{Without} & \textbf{Without} & \textbf{Without} & \textbf{Without} \\
 Micro-F1 & \textbf{EE} & \textbf{Coordinate} & \textbf{Sample} & \textbf{Missing} \\
\textcolor{red}{($\Delta$)} & \textbf{Training} & \textbf{Tokens} & \textbf{Strategy} & \textbf{Types} \\
\midrule
\multirow{1.75}[2]{*}{54.35} & 42.91 & 39.37 & 52.85 & 47.58 \\
 & \textcolor{red}{(-11.44)} & \textcolor{red}{(-14.98)} & \textcolor{red}{(-1.50)} & \textcolor{red}{(-6.77)} \\
\bottomrule
\end{tabular}%
    }}
\end{table}%

\clearpage{}
\vspace{-1mm}
\subsection{In-context Learning Performance}

\begin{figure}[t]
\begin{center}
\includegraphics[width=\linewidth]{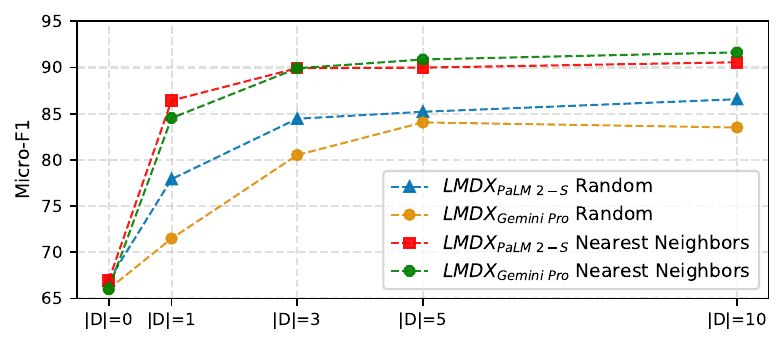}
\end{center}
\vspace{-5mm}
\caption{In-Context Learning results on CORD with random and nearest neighbors retrieval methods for LMDX\textsubscript{PaLM 2-S} and LMDX\textsubscript{Gemini Pro}.}
\label{fig:few-shot-cord}
\vspace{-3.5mm}
\end{figure}

In this section, we study how in-context learning (ICL) compares to finetuning. To do so, we test two methodologies: \textit{Random}, which randomly selects $|\mathcal{D}|$ documents and extractions from the train set, and \textit{Nearest Neighbors}, which uses similarity based on SentenceT5 embeddings \citep{ni2021sentence} to retrieve $|\mathcal{D}|$ documents to add in the LLM context. The results on CORD for LMDX\textsubscript{PaLM 2-S} and LMDX\textsubscript{Gemini Pro} are shown in Figure~\ref{fig:few-shot-cord}. Overall, while both methods increase the performance significantly, nearest neighbors shows a clear advantage, matching the best random ICL performance with only a single in-context example (86.43\% versus 86.57\% micro-F1 for LMDX\textsubscript{PaLM 2-S}), and matching the finetuned performance at $|\mathcal{D}|=10$ examples (90.33\% versus 90.02\% micro-F1), as examples from the same template are retrieved (see Appendix~\ref{appendix:icl_nearest_neighbors}). Beyond $|\mathcal{D}|=10$, the quality plateaus as no more example fit in the prompt.

\vspace{-1mm}
\section{Conclusion}
\vspace{-1mm}

In this paper, we have introduced LMDX, a methodology that enables using LLMs for information extraction on visually rich documents. With its coordinate tokens and decoding strategy, LMDX allows the high-quality extraction of singular, repeated and hierarchical entities, while localizing the entities in the document. LMDX is data efficient, and even allows extraction at zero-shot on entirely new document types and schemas. LMDX can benefit from orthogonal research fields, and we continue the discussion in the Limitations section. %

\vspace{-1mm}
\section*{Acknowledgements} 
\vspace{-1mm}

The authors would like to thank Tania Bedrax-Weiss, Riham Mansour, Slav Petrov, Yunhsuan Sung, Mike Kwong and Chun-Liang Li for their valuable feedback on the experiments and paper. The authors also thank Nikolai Glushnev for discussions, along with the help naming LMDX.

\vspace{-1mm}
\section*{Limitations} 
\vspace{-1mm}

We acknowledge the limitations of LMDX from the following aspects to inspire future research in the field of information extraction and localization.

Firstly, LMDX's input is text lines and their bounding boxes, usually coming from OCR. This means that LMDX can not extract non-textual entities thus would not be able to extract an entity that would be an image embedded in a document (e.g. a product\_image entity in a product webpage). This also limits performance in high-data scenarios, as all page image information is discarded. Furthermore, such input means that LMDX is sensitive to errors from the OCR process (wrong reading order, incorrect line grouping, undetected text and erroneously recognized characters). Qualitatively, we have found that a common error type for LMDX is caused by OCR grouping multiple semantically different segments together (we give a deeper analysis with concrete examples in Appendix~\ref{appendix:error_analysis}). Thus, techniques aiming to improve this error type would be a worthwhile future research direction.

Moreover, LMDX's localization mechanism is applied at the line level, where we verify that the predicted text is indeed present on the line. If the entity text appears multiple times on the line, we don't have a definitive way to choose the correct text. Thus, LMDX's localization and bounding boxes are not reliable beyond line-level granularity. While sufficient for greatly speeding up human-in-the-loop interactions like prediction auditing/review, getting entity bounding boxes precise at character-level would be beneficial, more natural-looking, and a worthy research direction.

Lastly, LMDX relies on LLMs supporting thousands of tokens in input and output (as detailed in Appendix~\ref{appendix:token-length-stats}), which is both computationally expensive and slow, requiring the use of hardware acceleration for acceptable latency and throughput. We showcase a latency comparison between popular solutions in Appendix~\ref{appendix:computational_cost}. General research in accelerating LLM inference \citep{shazeer2019fast, ainslie2023gqa, leviathan2023fast, hong2023flashdecoding++} would make LMDX more cost-effective in production setting. Specifically for LMDX, the coordinate tokens represent a large part of the total number of tokens, so research on minimizing their number (e.g. by introducing dedicated coordinate tokens within the LLM vocabulary) would yield significant savings and be a worthwhile direction for future work.

\bibliography{main}

\clearpage
\appendix

\section{Appendix}

\subsection{Chunking algorithm}
\label{appendix:chunking}

\begin{minipage}{\textwidth}
\begin{algorithm}[H]
  \footnotesize
   \caption{Document Chunking}
    \label{algo:chunking}
    \begin{algorithmic}[1]
    
    \Function{CHUNK}{$D$, $L$, $F$}
    \Comment{$D$ is a document containing multiple pages. $L$ is token limit.}
    \State \Comment{$F$ is a function that outputs prompt token length given some segments (e.g. lines).}
        \State $C = \phi$ \Comment{$C$ is to record all produced chunks.}
        \For{$i=1$ to $\lvert D.pages \rvert$}
            \State $S = D.pages[i].segments$
            \While{$S \ne \phi$}
                \For{$j=\lvert S \rvert $ to $1$} \Comment Start pruning from the end of the page.
                    \If{$F(S[1:j]) \le L$}
                        \State $C = C \cup  \{S[1:j]\}$
                        \State $S = S[j+1:\lvert S \rvert ]$ \Comment{Continue to work on the rest of the segments.}
                        \State Exit for loop
                    \EndIf
                \EndFor
            \EndWhile
        \EndFor
        \State \Return $C$
    \EndFunction
\end{algorithmic}
\end{algorithm}
\end{minipage}

\subsection{Entity value parsing algorithm}
\label{appendix:value_parsing}

\begin{minipage}{\textwidth}
\begin{algorithm}[H]
  \footnotesize
   \caption{Entity Value Parsing}
    \begin{algorithmic}[1]
    \Function{ParseEntityValue}{$D$, $E$} \Comment{$D$ is a document chunk.}
    \State \Comment{$E$ is raw extraction results for one entity type parsed from one LLM sample.}
        \State $G = \phi$ \Comment{$G$ is to record all parsed entity values.}
        \State $R = Regex(``(\symbol{92}d\symbol{92}d\symbol{92}|\symbol{92}d\symbol{92}d)")$ \Comment{$R$ is a regex that captures the segment identifiers.}
        \State $M = \{``s.x|s.y" \mapsto s | s \in D.segments\}$ \Comment{$M$ holds a mapping between segment id and segment.}
        
        \For{$i$ = 1 to $\lvert E \rvert$}
            \State $W = \phi$ \Comment{$W$ is to hold all words for this entity.}
            \State $P$ = $R.split(E[i])$ \Comment{P is expected to be interleaved text values and segment ids.}
            \For{$j$ = 1 to $\lvert P \rvert / 2$}
                \If{$P[j * 2] \notin M$}
                    \State Go to next $i$ \Comment{Segment ID is hallucinated. Grounding failure.}
                \EndIf
                \State $S = M[P[j * 2]]$ \Comment{Retrieve the stored segment from $M$ with parsed segment ID.}
                \State $T = P[j * 2 - 1]$ \Comment{$T$ is to hold the predicted text.}
                \If{$T$ not substring of $S$}
                    \State Go to next $i$ \Comment{Grounding failure, skip the current entity.}
                \EndIf
                    \State $W = W \cup (S \cap T)$
            \EndFor
            \State $G^\prime.value = \bigcup_{w \in W} w.text \_ value$ \Comment $G^\prime$ is to hold the entity to return.
            \State $G^\prime.bounding \_ box = \{\min(b.x), \min(b.y), \max(b.x), \max(b.y)\}_{w \in W, b=w.bounding \_ box}$
            \State $G = G \cup \{G^\prime\} $
        \EndFor
        
        \State \Return $G$
    \EndFunction
\end{algorithmic}
\end{algorithm}
\end{minipage}

\clearpage

\subsection{Decoding algorithm}
\label{appendix:decoding}

\begin{minipage}{\textwidth}
\begin{algorithm}[H]
  \footnotesize
   \caption{Responses Decoding}
    \begin{algorithmic}[1]
    \Function{DecodeForType}{$J$, $T$, $D$} \Comment{$J$ is one or more JSON objects.}
    \State \Comment{$T$ is an entity type.}
    \State \Comment{$D$ is a document chunk.}
        \State $E = \phi $ \Comment{$E$ is to record all parsed and grounded entities.}
        \For{$j$ = 1 to $\lvert J \rvert$}
            \State $J^\prime = J[j][T.type]$ \Comment{$J^\prime$ is to hold entities for T's type before grounding.}
            \If{$T.subtypes = \phi$} \Comment{$T$ is leaf entity type.}
                \State $E = E \cup ParseEntityValue(D, J^\prime)$
            \Else \Comment{$T$ is hierarchical entity type.}
                \State $E^\prime.subtypes = \bigcup_{T^\prime \in T.subtypes} DecodeForType(J^\prime, T^\prime, D)$ \Comment{$E^\prime$ is hierarchical entity.}
                \State $E = E \cup \{E^\prime\}$
            \EndIf
        \EndFor
        \State \Return $E$
    \EndFunction
    \State
    \Function{MajorityVoting}{$T$, $E$}\Comment{$T$ is an entity type.}
    \State \Comment{$E$ is a 2D vector of entities of type $T$ from all LLM responses.}
        \State $V = [0, 0, ..., 0] \in \mathbb{R} \textsuperscript{$\lvert E \rvert$}$ \Comment{$V$ is to record all votes.}
        \State $L = \{T\}$
        \While{$L \ne \phi$}
            \State $T^\prime = L[0]$
            \State $E^\prime = \phi$
            \For{$j$ = 1 to $\lvert E \rvert$}
                \State $E^\prime = E^\prime \cup \{e | e \in E[j], e.type = T^\prime\}$ 
                \Comment{$E^\prime[j]$ holds entities with type $T^\prime$ from E[j].}
            \EndFor
            \For{$i$ = 1 to $\lvert E^\prime \rvert$ - 1}
                \For{$j$ = $i + 1$ to $\lvert E^\prime \rvert$}
                    \If{$E^\prime[i] = E^\prime[j]$}
                        \State $V[i] = V[i] + 1$
                        \State $V[j] = V[j] + 1$
                    \EndIf
                \EndFor
            \EndFor
            \State $L = L[1:\lvert L \rvert]$ \Comment{Remove $T^\prime$ and inject its sub-types for recursion.}
            \State $L = L \cup T^\prime.subtypes$
        \EndWhile
        \State \Return $E[argmax(V)]$ \Comment{Return the entity values with the highest votes.}
    \EndFunction
    \State
    \Function{DecodeAllSamples}{$S$, $T$, $D$} 
    \Comment{$S$ is all LLM response samples on $D$.}
    \State \Comment{$T$ is a list of entity types.}
    \State \Comment{$D$ is a document chunk.}
        \State \Return $\bigcup_{T^\prime \in T} MajorityVoting(\bigcup_{S^\prime \in S} DecodeForType(ParseJson(S^\prime), T^\prime, D))$
    \EndFunction
\end{algorithmic}
\end{algorithm}
\end{minipage}

\clearpage
\subsection{Token Length Statistics}
\label{appendix:token-length-stats}

Table~\ref{table:io-token-length} details the token length (50\textsuperscript{th} and 99\textsuperscript{th} percentiles) of the prompt and completion targets for LMDX\textsubscript{PaLM 2-S} for the train split of datasets used in our experiments. We select the line level segment, 2 coordinate scheme, no JSON indentation so that all datasets fit within our 6144 prompt token length and 2048 output token length.

\begin{minipage}{\textwidth}
\begin{table}[H]
\caption{Prompt and target token length of different coordinate-as-tokens schemes on VRDU and CORD benchmarks, using the vocabulary of PaLM 2-S. We vary the number of coordinates and their quantization buckets in the localization tags, the segment level (e.g. line versus word), chunking style (e.g. page versus max input tokens) and JSON indentation in the schema and completion targets.}
\label{table:io-token-length}
  \centering
  \small
  \resizebox{\linewidth}{!}{
    \setlength{\tabcolsep}{3mm}{
\begin{tabular}{ccccccccc}
\toprule
 \multicolumn{9}{c}{} \\
 \multicolumn{9}{c}{\textbf{VRDU Ad-Buy Form}} \\
 \cmidrule(lr){1-9}

 \multirow{2}[2]{*}{\textbf{\# Coord.}}  &
 \multirow{2}[2]{*}{\textbf{\# Quant.}} &
 \multirow{2}[2]{*}{\textbf{Segment}} &
 \multirow{2}[2]{*}{\textbf{Chunking}} &
 \multirow{2}[2]{*}{\textbf{JSON Indent}} &
 \multicolumn{2}{c}{\textbf{Input}} &
 \multicolumn{2}{c}{\textbf{Target}} \\
\cmidrule(lr){6-7}
\cmidrule(lr){8-9}
 & & & & & \textbf{50\textsuperscript{th}} & \textbf{99\textsuperscript{th}} & \textbf{50\textsuperscript{th}} & \textbf{99\textsuperscript{th}} \\
\cmidrule(lr){1-9}
2 & 100 & Line & Page & None & 2377 & 3920 & 602 & 1916 \\
2 & 100 & Word & Page & None & 3865 & 13978 & 718 & 2328 \\
4 & 100 & Line & Page & None & 3329 & 5284 & 777 & 2473 \\
2 & 1000 & Line & Page & None & 2687 & 4322 & 660 & 2095 \\
2 & 100 & Line & Page & 4 & 2417 & 3328 & 689 & 2234 \\
2 & 100 & Line & 6144 tokens & None & 2377 & 3920 & 602 & 1916 \\
\toprule
 \multicolumn{9}{c}{} \\
 \multicolumn{9}{c}{\textbf{VRDU Registration Form}} \\
 \cmidrule(lr){1-9}

 \multirow{2}[2]{*}{\textbf{\# Coord.}}  &
 \multirow{2}[2]{*}{\textbf{\# Quant.}} &
 \multirow{2}[2]{*}{\textbf{Segment}} &
 \multirow{2}[2]{*}{\textbf{Chunking}} &
 \multirow{2}[2]{*}{\textbf{JSON Indent}} &
 \multicolumn{2}{c}{\textbf{Input}} &
 \multicolumn{2}{c}{\textbf{Target}} \\
\cmidrule(lr){6-7}
\cmidrule(lr){8-9}
 & & & & & \textbf{50\textsuperscript{th}} & \textbf{99\textsuperscript{th}} & \textbf{50\textsuperscript{th}} & \textbf{99\textsuperscript{th}} \\
\cmidrule(lr){1-9}
2 & 100 & Line & Page & None & 963 & 1578 & 79 & 147 \\
2 & 100 & Word & Page & None & 3083 & 5196 & 101 & 349 \\
4 & 100 & Line & Page & None & 1232 & 2017 & 91 & 177 \\
2 & 1000 & Line & Page & None & 1052 & 1723 & 83 & 155 \\
2 & 100 & Line & Page & 4 & 977 & 1592 & 92 & 160 \\
2 & 100 & Line & 6144 tokens & None & 963 & 1578 & 79 & 147 \\
\toprule
 \multicolumn{9}{c}{} \\
 \multicolumn{9}{c}{\textbf{CORD}} \\
 \cmidrule(lr){1-9}

 \multirow{2}[2]{*}{\textbf{\# Coord.}}  &
 \multirow{2}[2]{*}{\textbf{\# Quant.}} &
 \multirow{2}[2]{*}{\textbf{Segment}} &
 \multirow{2}[2]{*}{\textbf{Chunking}} &
 \multirow{2}[2]{*}{\textbf{JSON Indent}} &
 \multicolumn{2}{c}{\textbf{Input}} &
 \multicolumn{2}{c}{\textbf{Target}} \\
\cmidrule(lr){6-7}
\cmidrule(lr){8-9}
 & & & & & \textbf{50\textsuperscript{th}} & \textbf{99\textsuperscript{th}} & \textbf{50\textsuperscript{th}} & \textbf{99\textsuperscript{th}} \\
\cmidrule(lr){1-9}
2 & 100 & Line & Page & None & 342 & 869 & 355 & 1495 \\
2 & 100 & Word & Page & None & 396 & 1067 & 375 & 1638 \\
4 & 100 & Line & Page & None & 408 & 1139 & 422 & 1801 \\
2 & 1000 & Line & Page & None & 364 & 959 & 376 & 1957 \\
2 & 100 & Line & Page & 4 & 411 & 938 & 474 & 1997 \\
2 & 100 & Line & 6144 tokens & None & 342 & 869 & 355 & 1495 \\
\bottomrule
\end{tabular}
}}
\end{table}

\end{minipage}

\vspace{4cm}

\subsection{Schemas}
\label{appendix:schemas}

In this section, we present the schemas used for the experiments of this paper. The schema for VRDU Ad-Buy Form, VRDU Registration Form, and CORD can be found in Figure~\ref{fig:vrdu_ad_buy_schema}, Figure~\ref{fig:vrdu_registration_schema} and  Figure~\ref{fig:cord_schema} respectively. 

\clearpage
\begin{figure*}[ht]
\begin{minted}[frame=single,
               framesep=3mm,
               tabsize=4]{python}
{
  "advertiser": "",
  "agency": "",
  "contract_num": "",
  "flight_from": "",
  "flight_to": "",
  "gross_amount": "",
  "line_item": [
    {
      "channel": "",
      "program_desc": "",
      "program_end_date": "",
      "program_start_date": "",
      "sub_amount": ""
    }
  ],
  "product": "",
  "property": "",
  "tv_address": ""
}
\end{minted}
\caption{VRDU Ad-Buy Form Schema.} 
\label{fig:vrdu_ad_buy_schema}
\end{figure*}

\begin{figure*}[ht]
\begin{minted}[frame=single,
               framesep=3mm,
               tabsize=4]{python}
{
  "file_date": "",
  "foreign_principle_name": "",
  "registrant_name": "",
  "registration_num": "",
  "signer_name": "",
  "signer_title": ""
}
\end{minted}
\caption{VRDU Registration Form Schema.} 
\label{fig:vrdu_registration_schema}
\end{figure*}

\begin{figure*}[ht]
\begin{minted}[frame=single,
               framesep=3mm,
               tabsize=4]{python}
{
  "line_item": [ # menu
    {
      "discount_price": "", # menu.discountprice
      "identifier": "", # menu.num
      "name": "", # menu.nm
      "other": "", # menu.etc
      "quantity": "", # menu.qty
      "sub_name": [], # menu.sub_nm
      "sub_price": [], # menu.sub_price
      "sub_quantity": [], # menu.sub_qty
      "subtotal_price": "", # menu.itemsubtotal
      "total_price": "", # menu.price
      "unit_price": "" # menu.unitprice
    }
  ],
  "subtotal": { # subtotal
    "discount_price": "", # subtotal.discount_price
    "other": [], # subtotal.etc
    "service_price": "", # subtotal.service_price
    "subtotal_price": [], # subtotal.subtotal_price
    "tax_price": [] # subtotal.tax_price
  },
  "total": { # total
    "cash_price": [], # total.cashprice
    "change_price": "", # total.changeprice
    "credit_card_price": "", # total.creditcardprice
    "emoney_price": "", # total.emoneyprice
    "line_item_quantity_count": "", # total.menuqty_cnt
    "line_item_type_count": "", # total.menutype_cnt
    "other": "", # total.total_etc
    "total_price": [] # total.total_price
  }
}
\end{minted}
\caption{CORD Schema. Note that the original entity types (shown as comments) have been renamed to more semantically meaningful names.} 
\label{fig:cord_schema}
\end{figure*}

\clearpage

\subsection{Zero-shot Baselines Details}
\label{appendix:common_baseline}

We compare LMDX to other Large Model baselines on all benchmarks in the zero-shot context. Those baselines are detailed below.

\noindent \textbf{Text-based Baselines.} We evaluate the zero-shot extraction ability of multiple strong Large Language Models: GPT-3.5\footnote{https://platform.openai.com/docs/models/gpt-3-5-turbo}, Gemini Pro \citep{DBLP:journals/corr/abs-2312-11805} and PaLM 2-S \citep{palm2}. To do so, we prompt them with the raw benchmark's OCR text (no coordinate tokens or segment identifier like for LMDX), and extraction instructions alongside the schema in JSON format. We then parse the completions as JSON to get the predicted entities directly. Note that those predicted entities are not localized within the document. A sample prompt can be observed in Figure~\ref{fig:gpt35_sample}. In particular for GPT-3.5, we use the \texttt{gpt-3.5-turbo-1106} through OpenAI's API.

\noindent \textbf{Page Image-based Baselines.} We evaluate the zero-shot extraction ability of multiple strong Vision-Language Models: LLaVA-v1.5-13B \citep{liu2023improved}, Gemini Pro \citep{DBLP:journals/corr/abs-2312-11805}, and GPT-4V \citep{openai2023gpt4}. The prompt includes task description, instructions and target schema represented in JSON format as text input and the document page as image input. We provide examples of valid JSON values in the task instructions. Note that those predicted entities are not localized within the document. A sample prompt can be observed in Figure~\ref{fig:llava_sample}. In particular for GPT-4V, we use the \texttt{gpt-4-1106-preview} through OpenAI's API. For Gemini Pro, we use \texttt{gemini-pro-vision} through the VertexAI API\footnote{https://cloud.google.com/vertex-ai/docs/generative-ai/start/quickstarts/quickstart-multimodal}.

\subsection{CORD Baselines Details}
\label{appendix:cord_baseline}

\paragraph{LayoutLMv3 Baseline.} We follow the released implementation\footnote{https://github.com/microsoft/unilm/tree/master/layoutlmv3} for the $\rm{LayoutLMv3_{LARGE}}$ model and the training protocol described in \cite{huang2022layoutlmv3} as closely as possible. In particular, we train the model for $80$ epochs for each experiment on CORD (namely, $10$, $50$, $100$ and $200$-document training sets), on the IOB tags of the leaf entities. One difference in our training is that, due to computational resource constraints, we use $batch\_size = 8$ and $learning\_rate = 2\cdot10^{-5}$.

As the LayoutLMv3 model can only extract leaf entities, we design and heavily optimize a heuristic algorithm to group the leaf entities into hierarchical entities \textit{menu}, \textit{subtotal} and \textit{total}. The best heuristics we could find are as follows:
\begin{itemize}
    \item For the \textit{subtotal} and \textit{total} hierarchical entity types, since they appear only once per document, we group all their extracted sub-entities under a single \textit{subtotal} and \textit{total} entity, respectively. 
    \item For \textit{menu} hierarchical entity type, we observe that those entities usually occur multiple times on a document, and each \textit{menu} has at most one \textit{nm}, \textit{num}, \textit{unitprice}, \textit{cnt}, \textit{discountprice}, \textit{price}, \textit{itemsubtotal}, \textit{etc} sub-entities and potentially multiple \textit{sub\_nm}, \textit{sub\_price} and \textit{sub\_cnt} sub-entities. We also notice that the sub-entities aligned horizontally overwhelmingly belong to the same \textit{menu} entity, and a \textit{menu} entity can sometimes span over two or more consecutive horizontal lines. To leverage those observations, we perform a two-step grouping process for \textit{menu} entities. First, we merge the extracted leaf sub-entities into horizontal groups, where a threshold of $0.5$ on the intersection-over-union of the Y-axis was used for the determination of horizontal alignment. Second, we further merge the \emph{consecutive} horizontal groups into \textit{menu} entities, if and only if the horizontal groups do not have type duplication in any of the \textit{nm}, \textit{num}, \textit{unitprice}, \textit{cnt}, \textit{discountprice}, \textit{price}, \textit{itemsubtotal}, and \textit{etc} sub-entities (namely, those sub-entities only show up in at most one of the consecutive horizontal groups to be merged). We allow duplication of \textit{sub\_nm}, \textit{sub\_price} and \textit{sub\_cnt} sub-entity types. After those two steps, we obtain the final \textit{menu} entities.
\end{itemize}

\paragraph{Donut Baseline.} We follow Donut released implementation\footnote{https://github.com/clovaai/donut} for the Donut benchmarking results on CORD.
We use the default training configuration for all experiments on CORD (namely, $10$, $50$, $100$ and $200$-document training sets), with the following difference: we reduce batch size from 8 to 4 due to computational resource constraints,  and increase the number of train epochs from 30 to 60. For each experiment, checkpoint with the lowest loss on the dev set is selected and we report performance on test set. Micro-F1 scores produced by Donut evaluation code are reported (similar to all our other models).

\begin{figure*}[ht]
\begin{minted}[frame=single,
               framesep=3mm,
               fontsize=\small,
               tabsize=4]{text}
${RAW_OCR_TEXT}

Given the document, extract the text value of the entities included in 
the schema in json format.
- The extraction must respect the JSON schema.
- Only extract entities specified in the schema. Do not skip any entity types.
- The values must only include text found in the document.
- Use null or [] for missing entity types.
- Do not indent the json you produce.
- Examples of valid string value format: "$ 1234.50", "John Do", null.
- Examples of valid list value format: ["$ 1234.50", "John Do"], [].

Schema: {"file_date": "", "foreign_principle_name": "", 
"registrant_name": "", "registration_num": "", "signer_name": "",
"signer_title": ""}
```json
\end{minted}
\vspace{-3.25mm}
\caption{Sample prompt for GPT-3.5\textsubscript{+OCR}, Gemini Pro\textsubscript{+OCR} and PaLM 2-S\textsubscript{+OCR} text-based baselines for VRDU Registration Form.} 
\label{fig:gpt35_sample}
\end{figure*}

\begin{figure*}[ht]
\begin{minted}[frame=single,
               framesep=3mm,
               fontsize=\small,
               tabsize=4]{text}
${DOCUMENT_PAGE_IMAGE}

Given the document, extract the text value of the entities included in 
the schema in json format.
- The extraction must respect the JSON schema.
- Only extract entities specified in the schema. Do not skip any entity types.
- The values must only include text found in the document.
- Use null or [] for missing entity types.
- Do not indent the json you produce.
- Examples of valid string value format: "$ 1234.50", "John Do", null.
- Examples of valid list value format: ["$ 1234.50", "John Do"], [].

Schema: {"file_date": "", "foreign_principle_name": "", 
"registrant_name": "", "registration_num": "", "signer_name": "", 
"signer_title": ""}
```json
\end{minted}
\vspace{-3.25mm}
\caption{Sample prompt for GPT-4V\textsubscript{+Image}, LLaVA-v1.5-13B\textsubscript{+Image}, and Gemini Pro\textsubscript{+Image} image-based baselines for VRDU Registration Form.} 
\label{fig:llava_sample}
\end{figure*}

\subsection{Sample Prompts and Completions}
\label{appendix:sample_prompts_and_completions}

\begin{figure*}[!t]
\begin{center}
\includegraphics[width=0.98\linewidth]{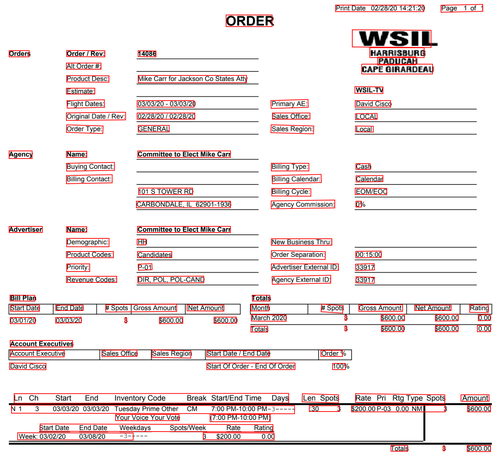}
\end{center}
\caption{Example document from VRDU Ad-Buy Form. The line bounding boxes detected by OCR are shown in red.}
\label{fig:vrdu_ad_buy_document}
\end{figure*}

In this section, we present example of LMDX prompts and completions from the LLM on the VRDU Ad-Buy dataset to better showcase the format used. Figure~\ref{fig:vrdu_ad_buy_document} shows the original document with the line bounding boxes from OCR, Figure~\ref{fig:vrdu_adbuy_sample} shows the corresponding prompt and completion on that document with coordinate segment identifiers, Figure~\ref{fig:vrdu_adbuy_sample_line_index} shows the same prompt and completion, but with line index segment identifiers (used in ablation studies to showcase how the LLM can interpret the layout) and Figure~\ref{fig:vrdu_adbuy_sample_missing_types} shows how the completion changes when skipping missing entity types entirely in the completion. Finally, Figure~\ref{fig:vrdu_adbuy_sample_4_line_coord} and Figure~\ref{fig:vrdu_adbuy_sample_2_word_coord} show how the prompts/completions changes if 4 line-level  $[x\textsubscript{min}, y\textsubscript{min}, x\textsubscript{max}, y\textsubscript{max}]$ coordinates or 2-word level $[x\textsubscript{center}, y\textsubscript{center}]$ coordinates are used.

\vfill

\subsection{Nearest Neighbors In-Context Learning}
\label{appendix:icl_nearest_neighbors}

\begin{figure*}[ht]
\begin{center}
\includegraphics[width=0.92\linewidth]{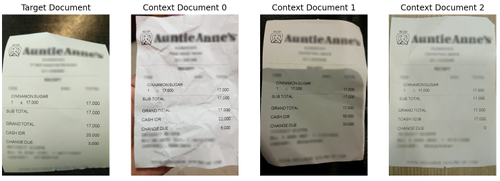}
\end{center}
\caption{Nearest Neighbors on CORD, Example 1, retrieving examplars from the same merchant.}
\label{fig:nn_1}
\end{figure*}

\begin{figure*}[ht]
\begin{center}
\includegraphics[width=0.92\linewidth]{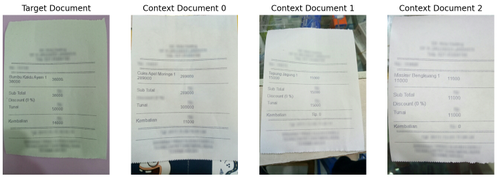}
\end{center}
\caption{Nearest Neighbors on CORD, Example 2, retrieving examplars from the same merchant.}
\label{fig:nn_2}
\end{figure*}

\begin{figure*}[ht]
\begin{center}
\includegraphics[width=0.92\linewidth]{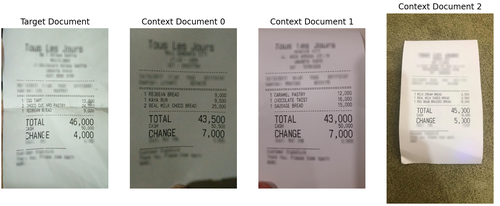}
\end{center}
\caption{Nearest Neighbors on CORD, Example 3, retrieving examplars from the same merchant.}
\label{fig:nn_3}
\end{figure*}

In our study, nearest neighbors leads to a significant quality gain over randomly selecting examplars. In this section, we explore why that is the case in the context of VRD information extraction. Figures~\ref{fig:nn_1}, \ref{fig:nn_2} and \ref{fig:nn_3} show typical retrievals using sentenceT5 \citep{ni2021sentence} embeddings\footnote{https://www.kaggle.com/models/google/sentence-t5/frameworks/tensorFlow2/variations/st5-base} on the OCR text for similarity. Unsurprisingly, nearest neighbors works well as it retrieves examplars from the same template as the target document, i.e. from the same merchant in the case of CORD documents (store/restaurant receipts). As those examples share the same layout, boilerplate text, and entities, it makes it easier for the model to understand the correct extraction pattern, leading to significant gains in quality.

\begin{figure*}[ht]
\begin{minted}[frame=single,
               framesep=3mm,
               fontsize=\small,
               tabsize=4]{text}
============================================= Prompt =============================================

<Document>
Print Date 02/28/20 14:21:20 75|02
Page 1 of 1 91|02
ORDER 49|04
WSIL 77|07
Orders 04|09
Order/Rev: 17|09
14086 29|09
HARRISBURG 78|09
PADUCAH 78|10
CAPE GIRARDEAU 78|12
Alt Order #: 17|11
Product Desc: 18|13
Mike Carr for Jackson Co States Atty 38|13
Estimate: 16|15
WSIL-TV 73|15
Flight Dates: 17|17
03/03/20 - 03/03/20 33|17
Primary AE: 57|17
David Cisco 73|17
Original Date/ Rev: 19|19
02/28/20/02/28/20 33|19
...
Start Date End Date Weekdays Spots/Week Rate Rating 31|66
Week: 03/02/20 03/08/20 -3-- 15|67
3 $200.00 0.00 47|67
Totals 79|69
3 87|69
$600.00 94|69
</Document>
<Task>
From the document, extract the text values and tags of the following 
entities:
{"advertiser": "", "agency": "", "contract_num": "", "flight_from": "", 
"flight_to": "", "gross_amount": "", "line_item": [{"channel": "", 
"program_desc": "", "program_end_date": "", "program_start_date": "", 
"sub_amount": ""}], "product": "", "property": "", "tv_address": ""}
</Task>
<Extraction>



=========================================== Completion ===========================================

{"advertiser": "Committee to Elect Mike Carr 37|36", "agency": 
"Committee to Elect Mike Carr 37|24", "contract_num": "14086 29|09", 
"flight_from": "03/03/20 33|17", "flight_to": "03/03/20 33|17", 
"gross_amount": "$600.00 77|51", "line_item": [{"channel": "3 31|63", 
"program_desc": "Tuesday Prime 31|63\nYour Voice Your Vote 29|64", 
"program_end_date": "03/03/20 31|63", "program_start_date": 
"03/03/20 31|63", "sub_amount": "$600.00 94|63"}], "product": 
"Mike Carr for Jackson Co States Atty 38|13", "property": "WSIL 77|07", 
"tv_address": null}
</Extraction>
\end{minted}
\caption{VRDU Ad-Buy Form sample prompt and completion with 2 $[x\textsubscript{center}, y\textsubscript{center}]$ coordinates for segment identifier. The document is truncated for easier visualization.} 
\label{fig:vrdu_adbuy_sample}
\end{figure*}

\begin{figure*}[ht]
\begin{minted}[frame=single,
               framesep=3mm,
               fontsize=\small,
               tabsize=4]{text}
============================================= Prompt =============================================

<Document>
Print Date 02/28/20 14:21:20 <0>
Page 1 of 1 <1>
ORDER <2>
WSIL <3>
Orders <4>
Order/Rev: <5>
14086 <6>
HARRISBURG <7>
PADUCAH <8>
CAPE GIRARDEAU <9>
Alt Order #: <10>
Product Desc: <11>
Mike Carr for Jackson Co States Atty <12>
Estimate: <13>
WSIL-TV <14>
Flight Dates: <15>
03/03/20 - 03/03/20 <16>
Primary AE: <17>
David Cisco <18>
Original Date/ Rev: <19>
02/28/20/02/28/20 <20>
...
Start Date End Date Weekdays Spots/Week Rate Rating <105>
Week: 03/02/20 03/08/20 -3-- <106>
3 $200.00 0.00 <107>
Totals <108>
3 <109>
$600.00 <110>
</Document>
<Task>
From the document, extract the text values and tags of the following 
entities:
{"advertiser": "", "agency": "", "contract_num": "", "flight_from": "", 
"flight_to": "", "gross_amount": "", "line_item": [{"channel": "", 
"program_desc": "", "program_end_date": "", "program_start_date": "", 
"sub_amount": ""}], "product": "", "property": "", "tv_address": ""}
</Task>
<Extraction>



=========================================== Completion ===========================================

{"advertiser": "Committee to Elect Mike Carr <44>", "agency": 
"Committee to Elect Mike Carr <29>", "contract_num": "14086 <6>",
"flight_from": "03/03/20 <16>", "flight_to": "03/03/20 <16>",
"gross_amount": "$600.00 <84>", "line_item": [{"channel": "3 <102>",
"program_desc": "Tuesday Prime Other CM <102>", "program_end_date": 
"03/03/20 <102>", "program_start_date": "03/03/20 <102>", 
"sub_amount": "$600.00 <100>"}], "product": "Mike Carr for Jackson Co 
States Atty <12>", "property": "WSIL-TV <14>", "tv_address": null}
</Extraction>
\end{minted}
\caption{VRDU Ad-Buy Form sample prompt and completion with line index for segment identifier. Instead of using coordinate tokens, the line index is given as \texttt{<line\_index>} at the end of each line, which does not communicate layout information but still uniquely identify the segment. The document is truncated for easier visualization.} 
\label{fig:vrdu_adbuy_sample_line_index}
\end{figure*}

\begin{figure*}[ht]
\begin{minted}[frame=single,
               framesep=3mm,
               fontsize=\small,
               tabsize=4]{text}
============================================= Prompt =============================================

<Document>
Print Date 02/28/20 14:21:20 75|02
Page 1 of 1 91|02
ORDER 49|04
WSIL 77|07
Orders 04|09
Order/Rev: 17|09
14086 29|09
HARRISBURG 78|09
PADUCAH 78|10
CAPE GIRARDEAU 78|12
Alt Order #: 17|11
Product Desc: 18|13
Mike Carr for Jackson Co States Atty 38|13
Estimate: 16|15
WSIL-TV 73|15
Flight Dates: 17|17
03/03/20 - 03/03/20 33|17
Primary AE: 57|17
David Cisco 73|17
Original Date/ Rev: 19|19
02/28/20/02/28/20 33|19
...
Start Date End Date Weekdays Spots/Week Rate Rating 31|66
Week: 03/02/20 03/08/20 -3-- 15|67
3 $200.00 0.00 47|67
Totals 79|69
3 87|69
$600.00 94|69
</Document>
<Task>
From the document, extract the text values and tags of the following 
entities:
{"advertiser": "", "agency": "", "contract_num": "", "flight_from": "", 
"flight_to": "", "gross_amount": "", "line_item": [{"channel": "", 
"program_desc": "", "program_end_date": "", "program_start_date": "", 
"sub_amount": ""}], "product": "", "property": "", "tv_address": ""}
</Task>
<Extraction>



=========================================== Completion ===========================================

{"advertiser": "Committee to Elect Mike Carr 37|36", "agency": 
"Committee to Elect Mike Carr 37|24", "contract_num": "14086 29|09", 
"flight_from": "03/03/20 33|17", "flight_to": "03/03/20 33|17", 
"gross_amount": "$600.00 77|51", "line_item": [{"channel": "3 31|63", 
"program_desc": "Tuesday Prime 31|63\nYour Voice Your Vote 29|64", 
"program_end_date": "03/03/20 31|63", "program_start_date": 
"03/03/20 31|63", "sub_amount": "$600.00 94|63"}], "product": 
"Mike Carr for Jackson Co States Atty 38|13", "property": "WSIL 77|07"}
</Extraction>
\end{minted}
\caption{VRDU Ad-Buy Form sample prompt and completion with skipping the missing entity types in the completion (tv\_address in this case). Instead of explicitly outputting the missing entity type \texttt{"tv\_address": null}, the entity type is skipped entirely. The document is truncated for easier visualization.}
\label{fig:vrdu_adbuy_sample_missing_types}
\end{figure*}

\begin{figure*}[ht]
\begin{minted}[frame=single,
               framesep=3mm,
               fontsize=\small,
               tabsize=4]{text}
============================================= Prompt =============================================

<Document>
Print Date 02/28/20 14:21:20 66|02|83|03
Page 1 of 1 87|02|95|03
ORDER 45|03|54|05
WSIL 69|06|85|08
Orders 02|09|06|10
Order/Rev: 14|09|21|10
14086 28|09|31|10
HARRISBURG 73|08|83|10
PADUCAH 74|10|82|11
CAPE GIRARDEAU 71|11|85|12
Alt Order #: 14|11|20|12
Product Desc: 14|12|22|13
Mike Carr for Jackson Co States Atty 28|12|49|14
Estimate: 14|14|19|15
WSIL-TV 70|14|75|15
Flight Dates: 14|16|21|17
03/03/20 - 03/03/20 27|16|39|17
Primary AE: 53|16|61|17
David Cisco 70|16|77|17
Original Date/ Rev: 14|18|25|19
02/28/20/02/28/20 27|18|39|19
...
Start Date End Date Weekdays Spots/Week Rate Rating 08|65|54|66
Week: 03/02/20 03/08/20 -3-- 04|66|27|67
3 $200.00 0.00 40|66|54|67
Totals 77|68|80|69
3 87|68|87|69
$600.00 92|68|96|69
</Document>
<Task>
From the document, extract the text values and tags of the following 
entities:
{"advertiser": "", "agency": "", "contract_num": "", "flight_from": "", 
"flight_to": "", "gross_amount": "", "line_item": [{"channel": "", 
"program_desc": "", "program_end_date": "", "program_start_date": "", 
"sub_amount": ""}], "product": "", "property": "", "tv_address": ""}
</Task>
<Extraction>



=========================================== Completion ===========================================

{"advertiser": "Committee to Elect Mike Carr 27|35|46|36", 
"agency": "Committee to Elect Mike Carr 28|24|46|25", "contract_num": "14086 28|09|31|10",
"flight_from": "03/03/20 - 03/03/20 27|16|39|17", 
"flight_to": "03/03/20 - 03/03/20 27|16|39|17", "gross_amount": null, 
"line_item": [{"channel": "N 1 3 0 ... Tuesday Prime Other CM 7:00 PM-10:00 PM-3----- 03|62|58|63", 
"program_desc": "Your Voice Your Vote 23|64|36|64", 
"program_end_date": "03/03/20 12|49|17|50", "program_start_date": "03/03/20 03|49|08|50",
"sub_amount": null}], "product": "WSIL-TV 70|14|75|15", "property": null, "tv_address": null}
</Extraction>
\end{minted}
\caption{VRDU Ad-Buy Form sample prompt and completion with 4 line-level $[x\textsubscript{min}, y\textsubscript{min}, x\textsubscript{max}, y\textsubscript{max}]$ coordinates in the prompt and completion. The document and completion is truncated for easier visualization.}
\label{fig:vrdu_adbuy_sample_4_line_coord}
\end{figure*}

\begin{figure*}[ht]
\begin{minted}[frame=single,
               framesep=3mm,
               fontsize=\small,
               tabsize=4]{text}
============================================= Prompt =============================================

<Document>
Print 67|02 Date 70|02 02/28/20 75|02 14:21:20 81|02 Page 88|02 1 91|02 of 93|02 1 95|02 
ORDER 49|04 WSIL 77|07 Orders 04|09 Order 15|09 / 18|09 Rev 19|09 : 21|09 14086 29|09
HARRISBURG 78|09 PADUCAH 78|10 CAPE 73|12 GIRARDEAU 80|12 Alt 14|11 Order 17|11 # 19|11 
: 20|11 Product 16|13 Desc 20|13 : 22|13 Mike 29|13 Carr 32|13 for 34|13 Jackson 38|13 
Co 41|13 States 45|13 Atty 48|13 Estimate 16|15 : 19|15 WSIL 71|15 - 74|15 TV 75|15 
Flight 15|17 Dates 19|17 : 21|17 03/03/20 30|17 - 33|17 03/03/20 36|17 Primary 56|17 
AE 59|17 : 61|17 David 71|17 Cisco 75|17 Original 16|19 Date 20|19 / 22|19 Rev 23|19 
: 25|19 02/28/20/02/28/20 33|19 Sales 55|19 Office 59|19 : 61|19 LOCAL 72|19 Order 15|21 
Type 19|21 : 21|21 GENERAL 31|20 Sales 55|21 Region 59|21 : 62|21 Local 72|21 Agency 05|24 
Name 15|24 : 17|24 Committee 31|24 to 35|24 Elect 38|24 Mike 41|24 Carr 44|24 Cash 71|26
Buying 16|26 Contact 20|26 : 23|26 Billing 15|28 Contact 20|28 : 22|28 Billing 55|26 
Type 59|26 : 61|26 Billing 55|28 Calendar 60|28 : 63|28 Calendar 73|28 101 29|30 S 30|30 
TOWER 34|30 RD 37|30 EOM 71|30 / 73|30 EOC 75|30 Billing 55|30 Cycle 59|30 : 61|30 
Agency 56|32 Commission 62|32 : 66|32 CARBONDALE 32|32 , 36|32 IL 37|32 62901-1936 42|32 
0 70|32 %
Mike 41|36 Carr 44|36 Demographic 17|38 : 22|38 HH 28|38 New 55|38 Business 59|38 
Thru 64|38 : 65|38 Product 16|39 Codes 20|39 : 23|39 Candidates 31|39 Order 55|39 
Separation 60|40 : 64|40 00:15:00 73|39 Priority 16|41 : 18|41 P 28|41 - 29|41 01 30|41
Advertiser 57|41 External 62|41 ID 65|41 : 67|41 33917 72|41 Revenue 16|43 Codes 21|43 : 
23|43 DIR 28|43 , 30|43 POL 32|43 , 33|43 POL 35|43 - 37|43 CAND 39|43 Agency 56|43
External 61|43 ID 64|43 : 65|43 33917 72|43 Bill 04|46 Plan 06|46 Start 04|48 Date 07|48 
End 13|47 Date 16|47 # 21|48 Spots 24|48 Gross 28|48 Amount 33|48 Net 38|47 Amount 42|48 
Totals 52|46 Month 51|47 March 51|49 2020 55|49 # 64|47 Spots 66|48 Net 83|47 Amount 86|47 
Gross 72|48 Amount 77|48 $ 75|49 600.00 78|49 Rating 94|48 0.00 95|49 03/01/20 05|49 
03/03/20 14|49 3 25|49 3 68|49 $ 32|49 600.00 34|49 $ 43|49 600.00 45|49 $ 86|49 
600.00 88|49 Totals 51|51 3 68|51 $ 75|51 600.00 77|51 $ 86|51 600.00 88|51 0.00 95|51 
Account 05|53 Executives 12|53 Account 05|54 Executive 10|54 Sales 22|54 Office 26|54 
Sales 32|54 Region 36|54 Start 42|54 Date 45|54 / 47|54 End 49|54 Date 52|54 Order 65|54 
Order 57|56 100 67|56 %
75|61 Rtg 78|61 Type 81|61 Spots 86|61 $ 69|63 200.00 72|63 P 74|63 - 75|63 03 76|63 
0.00 79|63 NM 82|63 3 87|63 Amount 93|61 $ 92|63 600.00 94|63 Ln 04|61 Ch 07|61 Start 13|61 
End 18|61 Inventory 26|61 Code 31|61 Break 39|61 Start 10|65 Date 13|65 End 17|66 Date 20|66
Weekdays 27|66 Spots 35|66 / 37|66 Week 39|66 Rate 46|66 Rating 52|66 Week 06|67 : 08|67 
03/02/20 11|67 03/08/20 18|67 -3-- 25|67 3 40|67 $ 43|67 200.00 45|67 0.00 53|67 Totals 79|69 
3 87|69 $ 92|69 600.00 94|69
</Document>
<Task>
From the document, extract the text values and tags of the following 
entities:
{"advertiser": "", "agency": "", "contract_num": "", "flight_from": "", 
"flight_to": "", "gross_amount": "", "line_item": [{"channel": "", 
"program_desc": "", "program_end_date": "", "program_start_date": "", 
"sub_amount": ""}], "product": "", "property": "", "tv_address": ""}
</Task>
<Extraction>



=========================================== Completion ===========================================

{"advertiser": "Committee to Elect Mike Carr 44|24", "agency": "Cash 71|26", 
"contract_num": "14086 29|09", "flight_from": "03/03/20 30|17", "flight_to": "03/03/20 36|17",
"gross_amount": "600.00 77|49", "line_item": [{"channel": "1 03|63", 
"program_desc": "Your Voice Your Vote 34|64", "program_end_date": "03/03/20 13|63",
"program_start_date": "03/03/20 19|63", "sub_amount": "600.00 92|63"}], 
"product": "WSIL 71|15 - 74|15 TV 75|15", "property": "WSIL 71|15 - 74|15 TV 75|15",
"tv_address": null}
</Extraction>
\end{minted}
\caption{VRDU Ad-Buy Form sample prompt and completion with 2 word-level $[x\textsubscript{center}, y\textsubscript{center}]$ coordinates in the prompt and completion. With word-level coordinate tokens, the text sequence of the prompt is mostly composed of coordinate tokens, and becomes far from the usual text sequences LLMs are trained on.}
\label{fig:vrdu_adbuy_sample_2_word_coord}
\end{figure*}

\clearpage
\clearpage %

\begin{table}[!th]
\caption{Parsing error rates from LMDX\textsubscript{PaLM 2-S} responses on VRDU Ad-Buy Mixed and CORD datasets.}
\label{table:sampling-errors}%
\centering
\resizebox{1.0\linewidth}{!}
{
    \setlength{\tabcolsep}{3mm}{
        \begin{tabular}{ccccc}
        \toprule
        \multirow{1.5}[2]{*}{\textbf{$|\mathcal{D}|$}}
        & \multirow{1.5}[2]{*}{Dataset}
        &  \multirow{1.5}[2]{*}{Invalid JSON}
        & Invalid Entity 
        & Entity Text \\
        & 
        & 
        & Value Format
        & Not Found \\
        \midrule
        \multirow{1.5}[2]{*}{\textbf{0}}
        & Ad-Buy & 0.18\% & 0.04\% & 0.59\% \\ 
        & CORD & 0.00\% & 0.00\% & 0.00\% \\
        \midrule
        \multirow{1.5}[2]{*}{\textbf{10}}
        & Ad-Buy & 0.27\% & 0.04\% & 0.44\% \\ 
        & CORD & 0.00\% & 0.00\% & 0.00\% \\
        \midrule
        \multirow{1.5}[2]{*}{\textbf{50}}
        & Ad-Buy & 0.24\% & 0.00\% & 0.17\% \\ 
        & CORD & 0.06\% & 0.00\% & 0.00\% \\
        \midrule
        \multirow{1.5}[2]{*}{\textbf{100}}
        & Ad-Buy & 0.24\% & 0.00\% & 0.13\% \\ 
        & CORD & 0.00\% & 0.03\% & 0.00\% \\
        \midrule
        \multirow{1.5}[2]{*}{\textbf{200}}
        & Ad-Buy & 0.25\% & 0.00\% & 0.09\% \\
        & CORD & 0.00\% & 0.00\% & 0.00\% \\
        \bottomrule
        \end{tabular}%
    }
}
\end{table}

\vspace{-5mm}
\subsection{Completion Parsing Error Rates}
\label{appendix:parsing_rates}

In this section, we report the various completion parsing error types and their occurrence rates for LMDX\textsubscript{PaLM 2-S}.

\noindent \textbf{Invalid JSON Formatting.} This refers to cases for which Python's \texttt{json.loads(completion)} fails on a LLM's completion. As observed in Table~\ref{table:sampling-errors}, the JSON parsing error rate is below 0.3\% in all training settings.

\noindent \textbf{Invalid Entity Value Format.} This refers to cases where the leaf entity value does not follow the expected \texttt{"<text-segment-1> XX|YY <text-segment-2> XX|YY"} format. As observed in Table~\ref{table:sampling-errors}, the Invalid Entity Value Format Rate is below 0.05\% in all training settings.

\noindent \textbf{Hallucination / Entity Text Not Found.} This refers to cases where the segment identifier is valid, but the entity text does not appear on the predicted segment (hallucination). As observed in Table~\ref{table:sampling-errors}, the Entity Text Not Found error rate is below 0.6\% in all training settings. As part of LMDX methodology, we discard any prediction whose text does not appear on the specified segment, ensuring we discard all hallucination.

Note that those numbers are computed at the completion level. Since multiple completions are sampled for each document chunk, the sampling scheme allows for correcting those errors and no document in the benchmarks fail extraction.

\subsection{Error Analysis}
\label{appendix:error_analysis}

\begin{figure}[!ht]
\begin{center}
\includegraphics[width=1.0\linewidth]{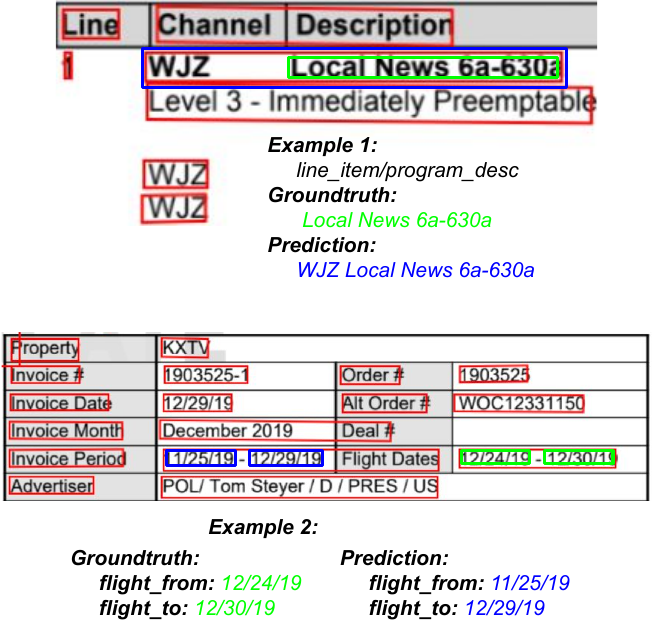}
\end{center}
\caption{Typical error pattern of LMDX\textsubscript{PaLM 2-S}. In both examples, the detected OCR lines are shown in red, the model predicted entities are shown in blue, and the groundtruth entities are shown in green. In both cases, the detected OCR lines merge two semantically distinct segments, causing the model to wrongly associate them in its predictions.}
\label{fig:error-analysis}
\end{figure}

In this section, we perform an error analysis on the test set to identify common error patterns of LMDX. A very common error type we observe is caused by OCR lines grouping multiple semantically different segments. We show two instance of those cases observed in LMDX\textsubscript{PaLM 2-S} on the VRDU Ad-Buy Form in Figure~\ref{fig:error-analysis}. In the first example, prediction for the entity \textit{line\textunderscore{item}/program\textunderscore{desc}} includes text from the previous column "Channel" along with the value in the column "Description". From the OCR line bounding boxes, we can see that these two columns are grouped as the same OCR line. In the second example, the model confuses between the adjacent keys "Invoice Period" and "Flight Dates" and extracts invoice dates as flight dates. Similar to the first example, OCR line bounding boxes show that the invoice dates and the key "Flight Dates" are grouped together in the same line although they are semantically different. As LMDX\textsubscript{PaLM 2-S} uses only coarse line layout information ([x\textsubscript{center}, y\textsubscript{center}] with 100 quantization buckets), the model fails in these cases, which is a current limitation of LMDX. 
We believe that incorporating the image modality will make LMDX more robust to those OCR errors.

\subsection{Effect of Coordinate Tokenization Schemes}
\label{appendix:coordinate_schemes}

In this section, we study how different designs of coordinate tokens affect extraction quality, to determine which is the most effective. There are many ways to tokenize the coordinates of the text segments, e.g. using line versus words as segment, the number of coordinates to use per segment ($[x\textsubscript{center}, y\textsubscript{center}]$ versus $[x\textsubscript{min}, y\textsubscript{min}, x\textsubscript{max}, y\textsubscript{max}]$). To establish which scheme is the most effective, we evaluate the zero-shot performance of LMDX\textsubscript{PaLM 2-S} on VRDU Registration Form, Ad-Buy Form and CORD benchmarks with the following schemes: 

\begin{itemize}
    \item 2 Line-level $[x\textsubscript{center}, y\textsubscript{center}]$ coordinates with $B = 100$ quantization buckets. 
    \item 4 Line-level $[x\textsubscript{min}, y\textsubscript{min}, x\textsubscript{max}, y\textsubscript{max}]$ coordinates with $B = 100$ quantization buckets. 
    \item 2 Word-level $[x\textsubscript{center}, y\textsubscript{center}]$ coordinates with $B = 100$ quantization buckets. 
\end{itemize}

\begin{figure}[ht]
\begin{center}
\includegraphics[width=\linewidth]{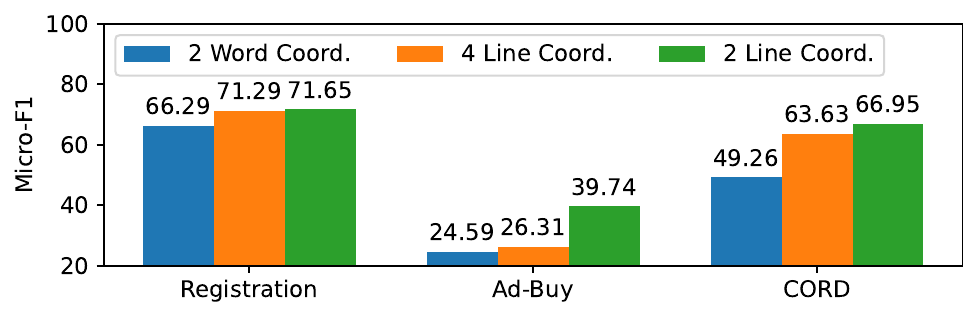}
\end{center}
\vspace{-3mm}
\caption{Zero-shot performance of LMDX\textsubscript{PaLM 2-S} on VRDU Registration Form Mixed Template, Ad-Buy Form Mixed Template, and CORD with different coordinate tokenization schemes. On all benchmarks, 2-line level $[x\textsubscript{center}, y\textsubscript{center}]$ coordinates obtains the best performance.}
\label{fig:coordinate_schemes}
\end{figure}

An example of prompt and completion for each scheme is given in Appendix~\ref{appendix:sample_prompts_and_completions}. Results quality is given in Figure~\ref{fig:coordinate_schemes}.

On all benchmarks, 2-line level $[x\textsubscript{center}, y\textsubscript{center}]$ coordinates obtains the best performance. This is caused by that coordinate tokenization scheme being very token efficient (see Appendix~\ref{appendix:token-length-stats}). 2 Word-level $[x\textsubscript{center}, y\textsubscript{center}]$ coordinates and 4 Line-level $[x\textsubscript{min}, y\textsubscript{min}, x\textsubscript{max}, y\textsubscript{max}]$ coordinates increase the tokens used by coordinates drastically, leading to multiple chunks generated for each document page, hence lowering quality since the model won't be able to correctly predict entities spanning multiple chunks. Furthermore, the large number of coordinate tokens added by those schemes makes the text sequences used in LMDX significantly different from the natural text sequences the LLM was pretrained on, hence making the LLM less effective at interpreting them, thereby lowering extraction quality.

\vspace{-4mm}
\subsection{Latency Comparison}
\label{appendix:computational_cost}

In Table~\ref{table:computational_cost}, we compare the latencies of LMDX and other methodologies, LayoutLMv3 and Donut. Overall, LayoutLMv3 is the fastest model, as it is encoder-only. LMDX\textsubscript{Gemini Pro} is the slowest among accelerated models, due to its reliance on an LLM with long inputs and outputs. Nonetheless, its latency remains acceptable in a production setting.

\begin{table}[h]
  \caption{Median, 95\textsuperscript{th} and 99\textsuperscript{th} percentile latencies of information extraction solutions on CORD. For text-based solutions LayoutLMv3 and LMDX\textsubscript{Gemini Pro}, the reported numbers do not include the OCR latency.}
  \label{table:computational_cost}%
  \centering
  \small
  \resizebox{\linewidth}{!}{
    \setlength{\tabcolsep}{2mm}
    {
\begin{tabular}{ccccc}
\toprule
 \multirow{2.25}[1]{*}{\textbf{Model}} & \multirow{2.25}[1]{*}{\textbf{Hardware}} & \multicolumn{3}{c}{\textbf{Latency (ms)}} \\
 \cmidrule(lr){3-5}
 &  & \textbf{Median} & \textbf{95\textsuperscript{th}\%} & \textbf{99\textsuperscript{th}\%} \\
\midrule
\multirow{1.75}[1]{*}{\textbf{LayoutLMv3}} & CPU & 648 & 999 & 1394 \\
\cmidrule{2-5}
 & GPU T4 & 30 & 55 & 93 \\
\midrule
\multirow{1.75}[1]{*}{\textbf{Donut}} & CPU & 14392 & 19427 & 20065 \\
\cmidrule{2-5}
 & GPU T4 & 620 & 1230 & 1683 \\
\midrule
\textbf{LMDX\textsubscript{Gemini Pro}} & TPU & 3653 & 7102 & 8345 \\
\bottomrule
\end{tabular}%
    }}
\end{table}%

\subsection{Ablation Detailed Metrics.}
\label{appendix:per-entity-f1}

In Table~\ref{table:per-entity-f1}, we present the per-entity F1 score for the coordinate ablations to highlight which entity most benefit from the introduction of layout information, and explain the large difference observed.

\begin{table}[!h]
\caption{Ablation of coordinates for LMDX\textsubscript{PaLM 2-S}, where per-entity F1 scores are shown. Adding coordinates greatly increase the extraction quality of entity types that require the understanding of spatial alignment, such as \textit{line\_item}.}
\label{table:per-entity-f1}
  \centering
  \small
  \resizebox{\linewidth}{!}{
    \setlength{\tabcolsep}{1mm}{
\begin{tabular}{ccc|c|c|c}
\toprule
\multirow{1.5}[2]{*}{Entity} & Hierar- & Number of & With & Without & \multirow{1.5}[2]{*}{$\Delta$} \\
  & chical? & occurrence & Coord. & Coord. & \\
\midrule
advertiser & \xmark & 635 & \textbf{95.90} & 88.98 & -6.92 \\
agency & \xmark & 283 & \textbf{73.05} & 71.84 & -1.21 \\
contract\_num & \xmark & 624 & \textbf{78.37} & 74.00 & -4.37 \\
flight\_from & \xmark & 540 & \textbf{67.74} & 63.63 & -4.11 \\
flight\_to & \xmark & 538 & \textbf{75.57} & 70.38 & -5.19 \\
gross\_amount & \xmark & 629 & \textbf{98.86} & 98.47 & -0.39 \\
product & \xmark & 607 & \textbf{87.86} & 75.33 & -12.53 \\
tv\_address & \xmark & 535 & \textbf{81.04} & 78.13 & -2.91 \\
property & \xmark & 595 & \textbf{77.75} & 74.33 & -3.42 \\
line\_item & \cmark & 9163 & \textbf{39.35} & 18.35 & -21.00 \\

\bottomrule
\end{tabular}%
}}
\end{table}%

Overall, performance is lower across the board without coordinates, as the VRDU Ad-Buy form benchmark consists entirely of invoices, a very layout-heavy document, where the alignment of the key segments (e.g. "Contract Number: ", "Total Due:", "Product:") and value segments (e.g. "123456", "\$1000", "Political Advertisement") matters a lot for the correct understanding of most entities. Thus, removing the layout modality will have a stark difference in performance. However, not all entities are affected in the same way:

\begin{itemize}
\item 8 out of 9 leaf entities see single digit drops, with gross\_amount affected the least. \textit{gross\_amount} can be derived mostly without looking at its key (e.g."Amount Due") and simply using cues commonly found in money amount entities (e.g. "\$", "USD", etc).
\item \textit{line\_item}, the hierarchical entity in Ad-Buy form, is affected the most. This is expected as \textit{line\_item}'s components (\textit{channel}, \textit{program\_desc}, \textit{program\_start\_date}, \textit{program\_end\_date}, \textit{sub\_amount}) are always visually arranged in tables, hence understanding horizontal and vertical alignments of the different text lines (which the coordinates provide) are critical for the correct grouping of those components into \textit{line\_items}.
\item Since \textit{line\_items} are by far the most common entity in VRDU Ad-Buy form (9163 occurrences), they have the most effect on the overall micro-F1.
\end{itemize}

\vspace{-2mm}

\subsection{Datasets Details}
\label{appendix:datasets_details}

As part of this work, we used the following datasets:

\begin{itemize}
    \item \textit{VRDU} \citep{10.1145/3580305.3599929} consists of two public visually-rich documents information extraction benchmarks: \textit{Registration Form}, which consists of 1915 documents recording foreign agents' activities with the US government requiring public disclosure, and \textit{Ad-Buy Form}, which consists of 641 signed receipts between a TV station and a political campaign group, requiring public disclosure to the Federal Communications Commission. All data is in English, and publicly available for research purposes.
    \item \textit{CORD} \citep{park2019cord} consists of 1000 Indonesian receipts from various shops and restaurants, on which the authors de-identified identifiable information. The data is a mix of Indonesian and English languages, and is publicly available for research purposes.
    \item \textit{Payment} \citep{majumder2020representation} is a dataset of 14,237 single-page invoices, annotated with  Each invoice is from a different vendor, so the documents do not share any common templates. All the data is in English.
    \item \textit{Synthetic Forms}. Finally, we generated synthetic training data using blank PDF form templates crawled from government websites, and filled with fully synthetic values using an internal tool. All generated data is in English. 
\end{itemize}

All extraction benchmarks we used are publicly available for research purposes, and we limited their usage to this research work. We also checked these datasets to ensure that no personally identifiable information is involved.

\vfill{}
\clearpage

\end{document}